\documentclass[APA,LATO2COL]{WileyNJDv5}

\usepackage{amsmath, amssymb, pifont}
\usepackage{algorithm}
\usepackage{subcaption}
\usepackage{algpseudocode}

\usepackage{overpic}
\usepackage{tikz}
\usepackage{xcolor}
\usepackage{longtable}
\usepackage{xcolor}
\usetikzlibrary{bayesnet}
\usepackage{paralist}
\usepackage{hyperref}
\usepackage{microtype}
\usepackage{xspace}
\usepackage{placeins}

\articletype{Research Article}%

\startpage{1}

\definecolor{rgb_turquoise}{RGB}{49, 182, 187}
\definecolor{rgb_red}{RGB}{199, 18, 44}
\definecolor{rgb_gray}{RGB}{124, 130, 135}

\def\ADAPT{ADAPT\@\xspace}

\newcommand{\new}[1]{\textcolor{black}{#1}}

\newcommand{\Fig}[1]{Figure~\ref{#1}}
\newcommand{\tcmark}{{\color{rgb_turquoise} \ding{51}}} 
\newcommand{\gcmark}{{\color{rgb_gray} \ding{51}}} 
\newcommand{\rxmark}{{\color{rgb_red} \ding{55}}}       

\def\ie{i.e.,\@\xspace}

\begin{document}

\title{ADAPT: An Autonomous Forklift for Construction Site Operation\footnote{This document is the result of the research project AWARD funded by the European H2020 program, No. 101006817.}}

\author[1]{Johannes Huemer$^{*,}$}
\author[1]{Markus Murschitz$^{*,}$}
\author[1]{Matthias Schörghuber}
\author[1]{Lukas Reisinger}
\author[1]{Thomas Kadiofsky}
\author[1]{Christoph Weidinger}
\author[1]{Mario Niedermeyer}
\author[1]{Benedikt Widy}
\author[1]{Marcel Zeilinger}
\author[1]{Csaba Beleznai}
\author[1]{Tobias Glück}
\author[1,2]{Andreas Kugi}
\author[1]{Patrik Zips$^{*,}$}

\address[1]{
    \orgdiv{Center for Vision, Automation and Control}, 
    \orgname{AIT Austrian Institute of Technology GmbH},
    \orgaddress{\state{Vienna}, \country{Austria}}
}
\address[2]{
    \orgdiv{Automation and Control Institute}, 
    \orgname{Technische Universität Wien},
    \orgaddress{\state{Vienna}, \country{Austria}}
}

\corres{Johannes Huemer ({johannes.huemer@ait.ac.at})}

\abstract[Abstract]{Efficient material logistics are critical in controlling costs and schedules in the construction industry. However, manual material handling remains prone to inefficiencies, delays, and safety risks. Autonomous forklifts offer a solution to streamline on-site logistics, reducing reliance on human operators and mitigating labor shortages. This paper presents the development and evaluation of ADAPT (Autonomous Dynamic All-terrain Pallet Transporter), an autonomous off-road forklift designed for construction environments. Unlike structured warehouses, construction sites pose significant challenges, including dynamic obstacles, unstructured terrain, and varying weather conditions. 
To address these challenges, our system integrates AI-driven perception with traditional approaches for decision making, planning, and control, enabling reliable operation in complex environments. We validate the system through extensive real-world testing, comparing its performance against an experienced human operator across various weather conditions. 
Our findings demonstrate that autonomous outdoor forklifts can operate near human-level performance, offering a viable path toward safer and more efficient construction logistics.}


\maketitle
\begingroup
  \def\thefootnote{*}
  \footnotetext{These authors contributed equally to this work.}%
\endgroup

\section{Introduction}

Streamlined logistics are critical to managing both costs and timelines in the construction industry. Studies indicate that materials represent more than half of the total construction costs and influence up to 80\% of the project schedules \cite{Caldas2015}.
A well-orchestrated on-site logistics, including resource delivery, storage and distribution \cite{Thomas2005}, ensures that resources are available when and where they are needed, reducing delays and costs \cite{Donyavi2009}. Numerous studies have identified inefficient on-site material management as a major contributor to project delays, see~\cite{Khursheed2024}. 

Digitalization and automation have the potential to revolutionize construction logistics. While digitalization allows optimal planning and tracking of materials, machinery automation ensures fluid and timely material flow operation on the construction site.
A multitude of reliable and flexible autonomous machines may be utilized to service the site in pull, push, and just-in-time deliveries. 
In particular, forklifts are easy to bring to construction sites. They are flexible in operation, which makes them well suited for ground-based material flow, lifting the workload of restricted cranes. 
When seamlessly integrated into the construction process through IT services, forklifts can greatly improve site efficiency by minimizing downtime, streamlined material handling, and preventing both equipment idleness and worker delays \cite{barbosa2013heijunka}. 
With the increasing shortage of skilled workers, the rough and dangerous working conditions, and the required increase in efficiency and decrease in errors, automation of these machines seems mandatory. 
\begin{figure}[hbt!]
    \centering
    \includegraphics[width=1\linewidth]{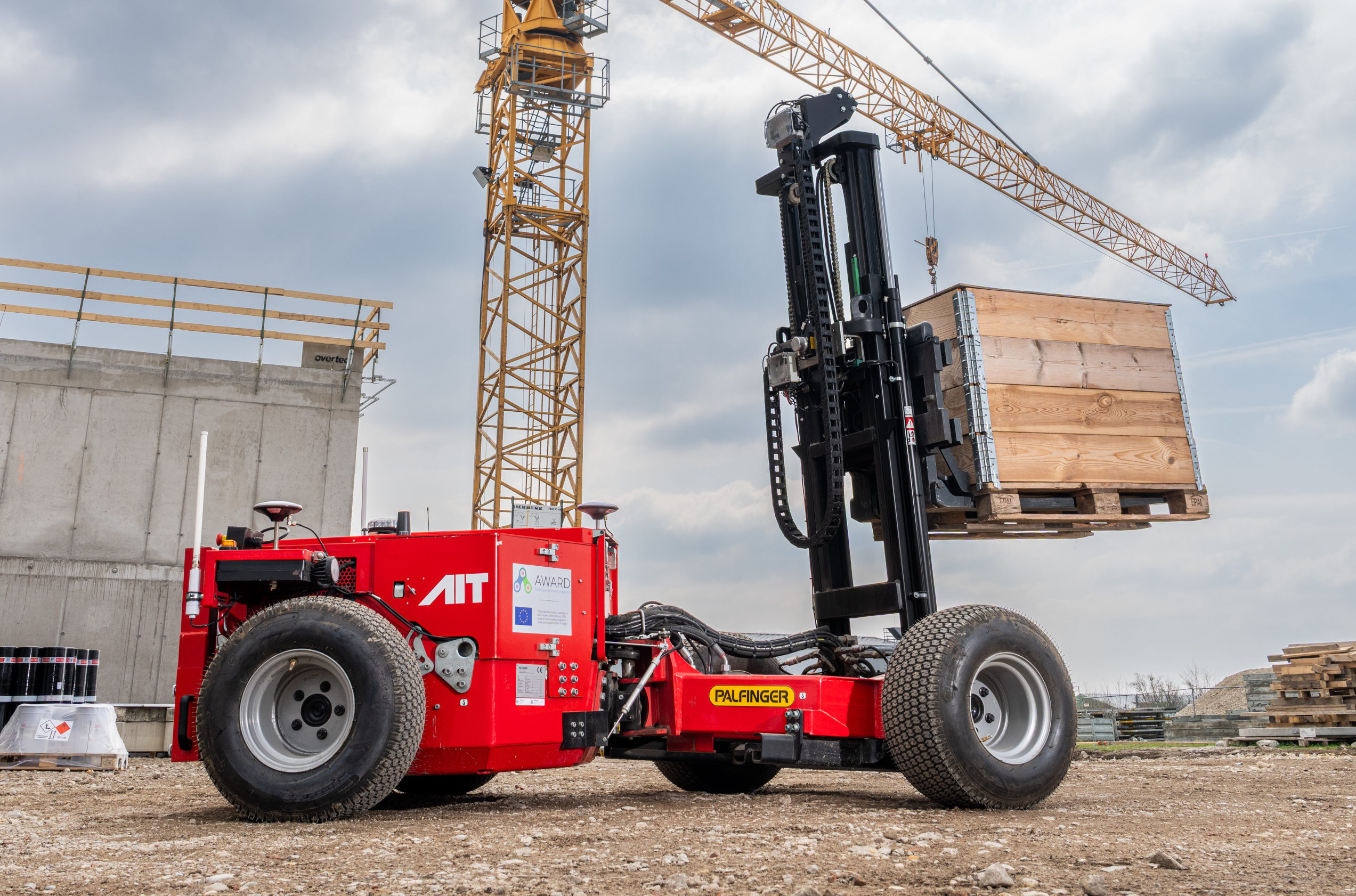}
    \caption{The autonomous off-road forklift ADAPT (Autonomous Dynamic All-terrain Pallet Transporter) in its designated working environment --- an unstructured construction site. 
    \newline Image credit: © AIT/tm-photography.}
    
     \label{fig:vehicle_in_action}
\end{figure}

However, the dynamic and unstructured nature of construction sites, combined with variable weather conditions pose significant challenges for automation. Unlike the controlled settings of autonomous warehouses, construction sites require flexible route planning and operation to navigate unpredictable terrains and obstacles. Safety is another paramount aspect, as 6\% of total construction costs are due to accidents, and one-third of all fatalities at construction sites are caused by material handling equipment \cite{Neitzel2001}.
\new{Despite substantial research on automation for construction, from individual autonomous machine to coordinated, on-site building \cite{Melenbrink2020}, large-scale outdoor autonomy is considerably more mature in other areas like surface mining, where autonomous haulage fleets have already moved billions of tonnes of material in routine commercial operation, with documented gains in productivity and safety \cite{CAT2024, Komatsu2024}, and agriculture, where autonomous platforms have reached large‐scale field deployment for operations such as harvesting, seeding, and spraying \cite{Ren2023}.}
Within construction itself, autonomous machines capable of demanding tasks such as loading and unloading full pallets from trucks remain scarce. This gap underscores the need for innovative systems that can seamlessly integrate into the demanding conditions of construction sites, a challenge widely recognized in recent reviews of robotics for construction that discuss the importance of control accuracy and condition monitoring \cite{Shi2023} and highlight limitations in perception, mobility, and safety integration \cite{liu2024}.

This paper presents the autonomous off-road forklift \ADAPT (Autonomous Dynamic All-terrain Pallet Transporter), depicted in Figure~\ref{fig:vehicle_in_action}. The scenario under consideration involves a truck-mounted forklift arriving on a typical construction site. That site is not designed for autonomous operation; predefined structures, guiding systems, or designated high-precision cargo locations are not required. Using advanced task and motion planning, synthetically trained geometry-based pallet detection, collision avoidance, and a new factor-graph-based joint vehicle localization and pallet mapping approach, our system aims to provide a reliable alternative to manual operation in this environment. 
The vehicle localization and pallet mapping approach is a variant of Simultaneous Localization and Mapping (SLAM) tailored to manipulation tasks. Unlike most SLAM techniques, which rely on natural features (mid-level) or raw sensor data (low-level), this approach operates at a higher level by utilizing object poses-specifically, pallet locations.
The overall approach is designed to ensure mandatory adaptability to dynamic construction site conditions while maintaining robust and reliable operation. This is achieved through a combination of novel AI-based approaches (e.g., pallet detection) and classical physics-driven methodologies for task management, motion planning, and control. 
Using primarily cost-effective sensors ensures affordability, while extensive real-world testing in various weather conditions validates the system’s performance. Through a comparative analysis with an experienced human operator, we demonstrate the potential of our autonomous forklift to give a realistic estimate of the performance and viability of autonomous machines in the coming years.
Since 100\% operational reliability is not yet achievable for flexible use on construction sites, the forklift has been designed with a seamless human-machine interaction concept. Human workers, typically present on-site, can intervene when necessary for operational or safety reasons. After such interventions, autonomous operation resumes smoothly without disruption.

\subsection{Contribution}
This paper presents a comprehensive system overview for an autonomous forklift for flexible outdoor operation. This includes hardware modifications and complete software integration for perception, planning and control specialized to the needs of construction sites. The main contributions are:
\begin{compactitem}
\item Development of an autonomous forklift system capable of loading and unloading operations in unstructured outdoor environments, demonstrating near-human performance across varied weather conditions, particularly robust operation under low to medium rainfall.
    
\item Design and implementation of a novel factor-graph-based joint optimization framework for precise vehicle localization and pallet mapping, specifically engineered to enhance object manipulation accuracy in pallet loading tasks.
    
\item Introduction of an innovative fork contact measurement system utilizing pressure feedback, significantly enhancing manipulation robustness and operational safety while maintaining cost-effectiveness and implementation simplicity.
    
\item Comprehensive performance evaluation comparing the autonomous system against an expert human operator with over 20 years of experience, including detailed quantitative analysis of operational efficiency, system robustness, and the frequency and severity of required human interventions during extended autonomous operation.
\end{compactitem}

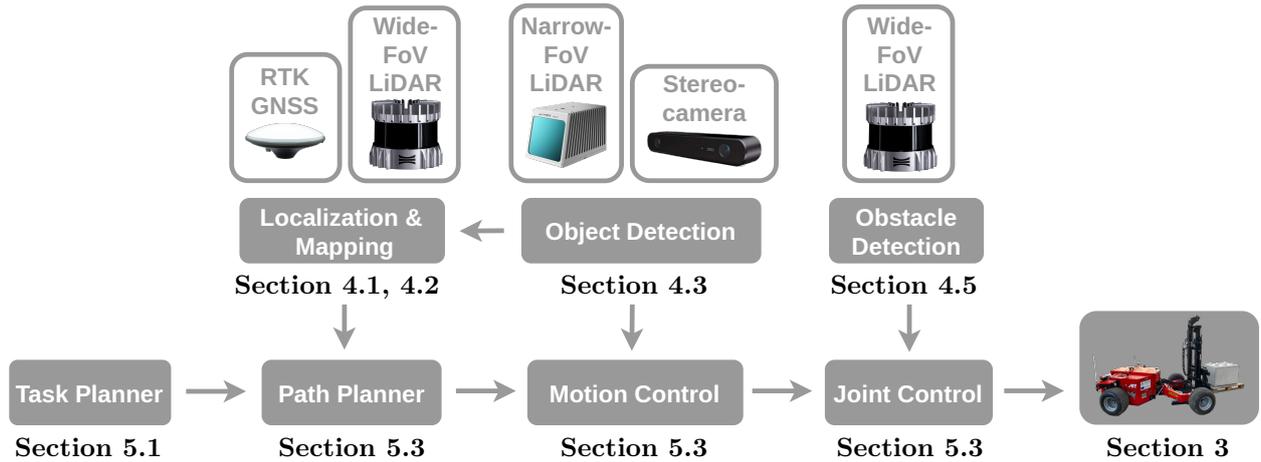
\begin{figure*}[ht!]
    \centering
    \begin{overpic}[width=0.9\linewidth]{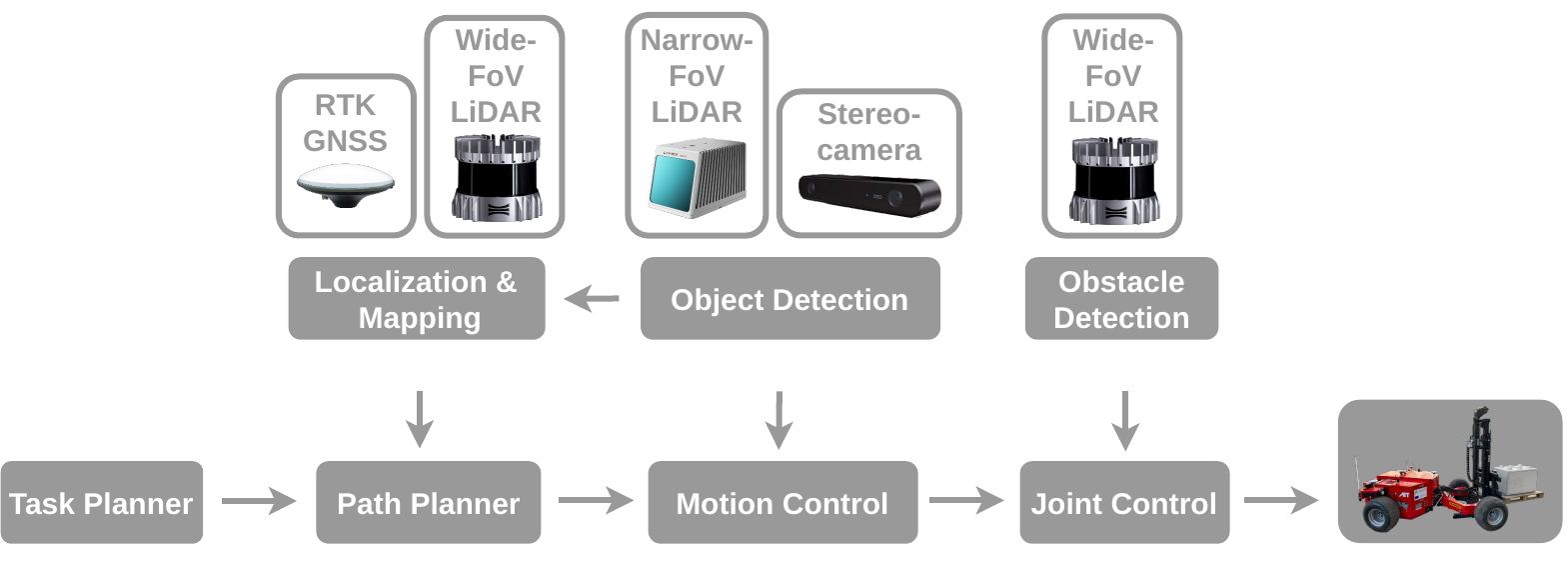}
        \put(1.5,-1){\textbf{Section~\ref{sec:task_planning}}}
        \put(22.5,-1){\textbf{Section~\ref{sec:motion_control}}}
        \put(45,-1){\textbf{Section~\ref{sec:motion_control}}}
        \put(67,-1){\textbf{Section~\ref{sec:motion_control}}}
        \put(88.5,-1){\textbf{Section~\ref{sec:system_description}}}
        \put(20,12){\textbf{Section~\ref{sec:state_estimation},~\ref{sec:mapping}}}
        \put(45,12){\textbf{Section~\ref{sec:pallet_detection}}}
        \put(67,12){\textbf{Section~\ref{sec:obstacle_detection}}}
    \end{overpic} 
    \vspace{0.2cm}
    \caption{System architecture of the proposed autonomous forklift, illustrating the hierarchical organization of key components essential for autonomous operation. The diagram depicts high-level information pathways and data flow between integrated sensors and specialized processing modules throughout the system. Relevant section references are provided alongside each major component, directing readers to corresponding detailed descriptions within the manuscript.}    
    \label{fig:component_overview}
\end{figure*}

\subsection{Article Structure}
The remainder of this paper is organized as follows. Section \ref{sec:related_work} reviews recent advances in autonomous outdoor machinery and related research in robotic material handling. Section \ref{sec:system_description} provides an overview of the proposed system, including its hardware and software components. The perception system, responsible for environment sensing, obstacle detection, and localization, is described in Section \ref{sec:perception}. Section \ref{sec:planning_control} details the planning and control strategies for autonomous navigation and load handling. In Section \ref{sec:experiments}, we evaluate the system's performance compared to an expert operator through experiments conducted in real-world outdoor environments and provide development and testing insights. Finally, Section \ref{sec:conclusion} concludes the paper and discusses potential directions for future work. An overview of the system components and the corresponding sections is sketched in \Fig{fig:component_overview}.

\section{Related Work}
\label{sec:related_work}
While automated manipulation and logistics are commercially viable in structured indoor environments, the deployment of autonomous machines in unstructured outdoor settings remains an active area of research. Nevertheless, recent years have produced several promising prototypes. The HEAP platform~\cite{Jud.2021} combines navigation and manipulation capabilities in a legged excavator. This outstanding research platform has demonstrated various tasks including excavation, dry-wall construction with natural stones \cite{Mascaro2021}, and forestry operations using both classical control techniques and novel AI approaches. 
\new{In a related effort, an autonomous excavator system \cite{Zhang2021} demonstrated long-duration, fully autonomous material loading in unstructured environments. The system reportedly achieved productivity comparable to that of an experienced human operator.} 
Similarly, the Harveri platform~\cite{Jelavic.2022} employs comparable technologies to harvest trees in challenging environments.

An automated logistics yard machine~\cite{Meyer2024} was introduced for mixed operations with both manual and automated vehicles. The machine's core capabilities—driving in open spaces and precise docking for trailer pick-up—closely resemble forklift operations. However, detailed technical information about the automation system remains unavailable.

Over the past two decades, several research platforms for autonomous forklift solutions have been published.
Earlier work~\cite{Seelinger2006} developed a prototype vision-guided forklift system that enabled precise pallet engagement using camera-space manipulation. Later research~\cite{Tews2007} explored vision-based handling tasks for autonomous outdoor forklifts, particularly focusing on vision systems for transporting molten aluminum in the metal industry. Another study~\cite{tamba2009path} presented a system configuration for autonomous forklift operation that incorporated vision, laser range finders, sonar, and other sensors. This research analyzed the kinematics of a spin-turn mechanism and established essential system equations for path-following based on time-varying feedback control law. However, in contrast to our work, their evaluation merely demonstrated the basic functionality of path following without presenting comprehensive real-world evaluations of loading success.

To our knowledge, only two similar scientific publications~\cite{Walter2015, Iinuma_2020} focus on autonomous forklifts in outdoor operation. The first~\cite{Walter2015} primarily emphasized motion planning with an anytime approach and explored new interaction modalities with human workers, such as voice commands. This work used trained AI only for object reacquisition after manual annotation of a single frame. Although it presented the robustness of the path planning, it did not provide a robustness or performance analysis of the complete loading cycle. Similarly, the second study~\cite{Iinuma_2020} concentrated on successfully unloading a specific pallet type from a truck, without offering comparative analysis against the performance of the human operator.

Another promising platform has been presented by the Linde group and their scientific collaborators, featuring counterbalanced forklifts capable of operating in both indoor and outdoor environments~\cite{Kanis2024}. This research emphasized cooperative behavior through real-time information exchange, handling of inclines and gradients, and management of weather influences. The work highlighted the necessity for enhanced performance in outdoor settings, including the ability to navigate inclines and adapt to varying weather conditions. However, no scientific publication detailing technical specifications or evaluation results is currently available.

\subsection{Localization, Mapping, and Traversability}

In order to enable a system to operate in unfamiliar environments, a common approach is to use variants of SLAM~\cite{dissanayake2001solution}. SLAM is a technique that simultaneously constructs a map of an unknown environment while tracking the agent's location within that environment, and has been widely studied and applied in robotics~\cite{cadena2016past, ebadi2024subt}. 
\new{Beyond structured indoor environments, localization and mapping have been extensively studied in autonomous driving, field robotics, and off-road autonomy. These domains share many challenges with autonomous outdoor forklift operation, including navigation in large-scale environments, varying environmental conditions, and heterogeneous sensor modalities. In \cite{min2026}, the authors survey the key challenges in off-road, unstructured environments and provide a comprehensive taxonomy of related work spanning both modular autonomous-function architectures and end-to-end methods.}

While mapping robot environments has been extensively studied over past decades, human-like scene understanding remains a key challenge for safe navigation in unstructured environments. Simple 2D occupancy grids may suffice for structured indoor spaces, but complex terrains require richer representations. The work \cite{hornung_octomap_2013} introduced a pioneering approach to address memory and runtime efficiency challenges in 3D occupancy maps, which has since been enhanced by methods such as spatio-temporal voxel layers~\cite{macenski_spatio-temporal_2020} and VDB (Volumetric Data Blocks)~\cite{besselmann_vdb-mapping_2021}. Other popular representations for 3D mapping include point clouds~\cite{krusi_driving_2017} and implicit representations~\cite{whelan_real-time_2015, oleynikova_voxblox_2017, vizzo_vdbfusion_2022}. 

\new{In addition to geometric reconstruction, autonomous operation in outdoor environments requires semantic scene understanding. Perception systems in autonomous driving and field robotics therefore integrate mapping with terrain classification, obstacle recognition, and traversability estimation to support planning and decision making. Recent surveys provide comprehensive overviews of these approaches. The work \cite{arafin2025} reviews advances in off-road terrain classification, focusing on sensor modalities, appearance- and geometry-based representations, and learning-based methods. Similarly, \cite{borges_survey_2022} surveys scene understanding and traversability analysis, covering both classical and deep learning approaches using proprioceptive and exteroceptive sensors, highlighting their effectiveness in enhancing navigation capabilities in diverse environments}.
When multiple sensors are used, it is common to implement certain aspects of sensor fusion and state estimation together with SLAM. For comparative overviews on sensor fusion techniques, see~\cite{yeong2021sensor}, and~\cite{ahangar2021survey} for a focus on autonomous driving applications.

An elegant unified approach is to use algorithms based on factor graphs~\cite{dellaert2017factor}. Although filter-based techniques perform well in state estimation~\cite{jung2021mmf, talbot2024continuoustimestateestimationmethods}, factor graphs have become a dominant methodology in modern SLAM systems~\cite{ebadi2024subt, nubert2022graph}. They allow for an efficient representation of both sensor fusion functionality and temporal dependencies between measurements. The frequency of these signals can vary in such systems, and algorithms exist that can solve for the most probable state of the system given a series of measurements, subject to the system's constraints (i.e., maximum a posteriori likelihood estimation). A notable algorithm for solving such problems is iSAM2~\cite{kaess2012isam2}, which uses factor graphs as its foundational framework while focusing on efficient incremental updates over time. 

\new{These developments have increasingly been adopted in autonomous material-handling systems. For autonomous forklifts, the SLAM method presented in \cite{WangFeiren2020} demonstrates robust operation in large-scale warehouse environments using stereo vision, by combining direct (pixel-based) and indirect (feature-based) SLAM techniques. However, such systems primarily target indoor environments and mostly static settings. In contrast, autonomous truck-mounted forklifts operating on construction sites face additional challenges arising from unstructured terrain, changing environmental conditions, dynamic obstacles, and the need for traversability-aware navigation.
Building upon these developments in SLAM, mapping, traversability analysis, and probabilistic state estimation, the proposed system employs a factor-graph formulation to jointly estimate both the poses of the forklift and the pallets within a unified optimization framework, combined with a 2.5D elevation map for traversability estimation.}


\subsection{Pallet Pose Estimation}
Automated pallet manipulation requires estimating the 3D pallet pose with respect to the vision sensor. As we focus on a specific pallet type of known size and geometry (Euro-pallet), the vision task becomes \textit{instance-level} pose estimation. Common pose-aware pallet detection approaches adopt geometric cues from depth data or rely on visual appearance. \textit{Geometric schemes} typically focus on 2D structure templates~\cite{Xiao2017} or perform geometric fitting on point-cloud data from stereo~\cite{Shao2023}, Time-of-Flight or LiDAR sensing~\cite{Iinuma_2020}. However, geometric analysis schemes inherently become ambiguous in the presence of clutter and partial occlusions. \textit{Appearance-based} neural representations, sometimes complemented by depth in the form of RGB-D images, yield highly object-specific detectors, as demonstrated for 3D pose estimation in recent challenges~\cite{BOP2023}. Especially occlusion-robust local key-point representations~\cite{vu2024, NVP} offer accurate pose estimation results in high-resolution and semantically rich RGB or \mbox{RGB-D} data spaces. 
The work \cite{kai2025pallet} introduces a pallet detection method based on the concept of \textit{Front Face Shots}, which represent the accessible side of pallets with the fork pockets. Their approach demonstrates that by combining a machine learning-based object detector with a kernel-based regression method, an accurate 6D pose can be computed, even for previously unseen pallet appearances.

Our employed pallet pose estimation scheme~\cite{beleznai_2024} exclusively relies on depth data to reduce object variations only to geometric traits. Furthermore, as a distinctive feature compared to most state-of-the-art, we only use stereo depth computed on synthetic image pairs for training, to generate an infinite diversity of view configurations with a narrow sim-to-real gap.

\subsection{Task Planning}
Task planning is essential for automating utility machinery, particularly in complex load manipulation tasks such as pallet handling.
It involves high-level decision-making and sequencing of operations.
Early approaches relied on finite state machines (FSMs) but evolved into hierarchical behavior trees for greater modularity and reactivity \cite{Ghzouli.2023}. 
Similarly, the autonomous system under evaluation employs a modular behavior tree implementation to efficiently manage task execution.
More recently, integrated Task and Motion Planning (TAMP) methods have emerged \cite{Garrett.2021}, combining symbolic reasoning with motion feasibility checks \cite{Hartmann.} or hierarchical architectures that leverage Linear Temporal Logic (LTL) for high-level planning and reactive behavior trees for low-level control \cite{Li.}. Learning-based methods are explored to infer planning domains from data, reducing the reliance on manually defined preconditions \cite{Huang.2025}.
Recent developments have deployed LLMs for task planning. Although LLMs excel at processing natural language and reasoning, directly translating abstract language inputs into executable plans is problematic due to their limited grounding in physical environments and inability to reason over complex task sequences \cite{birr2024autogpt+}.
LLMs struggle with the physical understanding of actions and fail to manage long-term dependencies in multiple steps, limiting their applicability to planning real-world tasks \cite{Guan2023}.
Recent research has explored the integration of LLMs with classical planners to capitalize on the language understanding of LLMs while leveraging the precision of PDDL-based planning \cite{Wang2023, Hu2024}. This hybrid approach improves the decision making capabilities of autonomous agents, allowing them to process natural language instructions and generate actionable plans using PDDL-based frameworks \cite{Wang2024a, Liu2023}. However, no reliable long-term operation is yet feasible.

\subsection{Motion Planning and Control}

Motion planning and control ensure task feasibility, forming an adaptive feedback loop for execution. In the context of vehicles, motion planning must account for non-holonomic constraints such as limited turning radii and restricted maneuverability~\cite{Sahoo.2023, NonHolonomic2006}. One such constraint arises from the articulation of the center, a common feature in heavy-duty vehicles, which significantly influences the kinematic behavior.

\new{Recent literature has addressed these challenges through comprehensive surveys that categorize hierarchical and end-to-end planning approaches and analyze terrain-dependent constraints and cost functions \cite{wang2024b}}. A Goal-Directed Rapidly exploring Random Tree (RRT) with multi-step refinement has been proposed for articulated construction machines~\cite{Tong.2019}. Similarly, an extended Reeds-Sheep algorithm has been applied to wheel loader path planning~\cite{Alshaer.2013}. A modified bug-like algorithm, integrated with Model Predictive Control (MPC) to ensure smoother trajectory execution, was proposed in~\cite{Nayl.2015}. Additionally, offline motion primitive generation and online receding-horizon planning have been successfully applied to tree harvester vehicles~\cite{Hu.}. Application-focused path planning for yard automation was proposed in~\cite{belov.2021}.

For tracking and control, a variant of the pure-pursuit method for articulated vehicles was introduced in~\cite{Hu.}. An alternative approach using iterative learning control in a feedback linearized space has been validated in large-scale field experiments with center-articulated mining vehicles~\cite{Dekker2019}. Another tracking controller was specifically developed for articulated drum rollers navigating construction sites~\cite{bian2017kinematics}.

Our navigation approach employs the well-established Hybrid A* planner with a Reeds-Shepp configuration~\cite{Dolgov.2010} by introducing an analytic conversion between car-like and articulated vehicles for constant curvature driving. This path planning is complemented by a Lyapunov-based path tracking controller similar to~\cite{bian2017kinematics}.

Beyond collision-free navigation, precise pallet docking is crucial for autonomous forklifts. Successful docking relies on accurate pallet pose estimation, achieved through vision-based systems, laser range finders, or hybrid sensor fusion approaches. The literature explores various motion planning techniques for this task, including geometric-based path generation~\cite{Iinuma_2020,Molter.} and dynamic-based planning methods~\cite{Baglivo.2011}. In addition, advanced control strategies, such as visual servoing~\cite{Tsiogas.}, enhanced the accuracy of jacking. Some methods also consider pallet inclination during insertion~\cite{Kita.}, improving adaptability and robustness in unstructured environments.

For the system presented in this paper, the docking approach is based on visual servoing, integrating depth-camera and LiDAR measurements to achieve precise pallet engagement under varied environmental conditions.

\section{System Design}
\label{sec:system_description}
This section describes ADAPT, including the vehicle platform it is based on, the hardware modifications necessary for automation, and an overview of the key software components.

\begin{figure*}[ht!]
    \centering
    \includegraphics[width=0.99\linewidth]{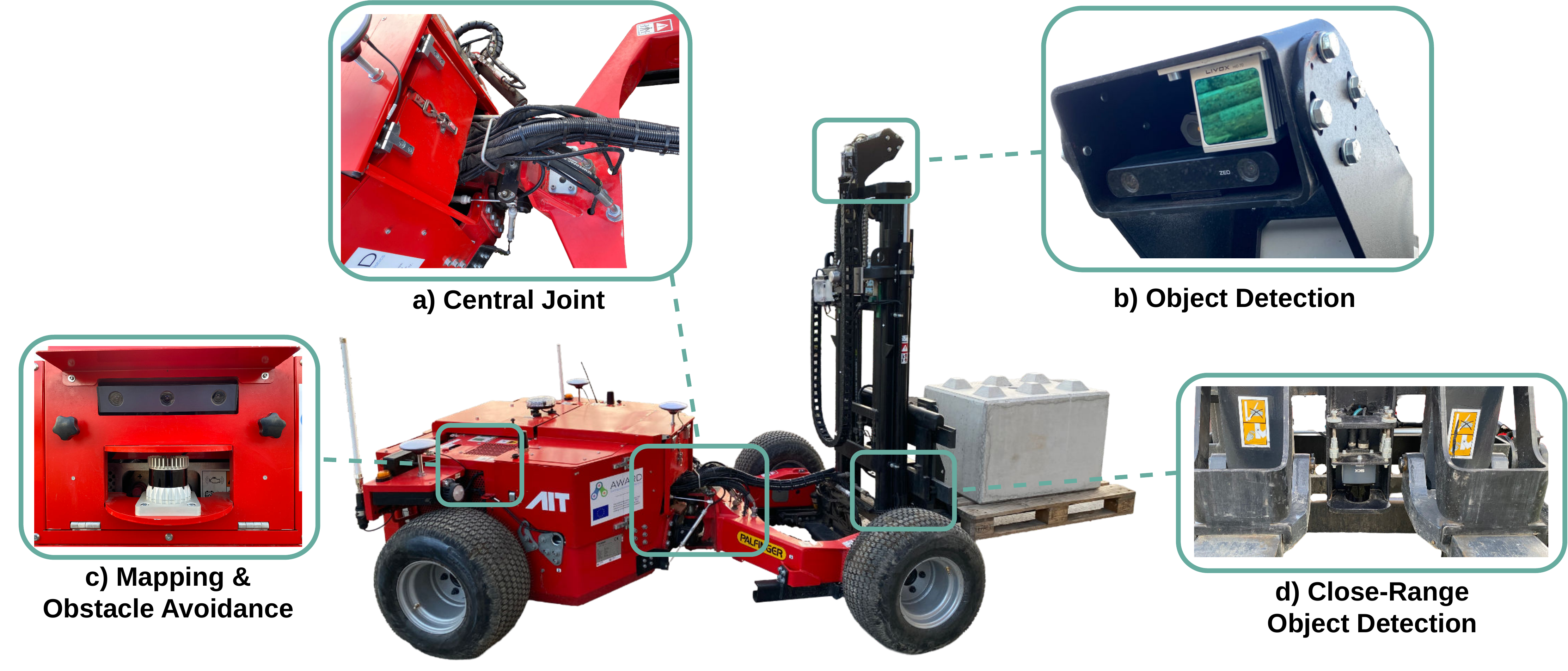}
    \vspace{-5pt}
    \caption{Key hardware components of ADAPT. (a) 3D central joint for maneuverability and structural flexibility, (b) Object detection system, (c) Terrain mapping and obstacle avoidance sensor, and (d) Close-range object detection for precise load handling.}
    \label{fig:system_overview}
\end{figure*}

\subsection{Vehicle Platform}

\begin{figure}[h]
    \centering
    \includegraphics[width=1.0\linewidth]{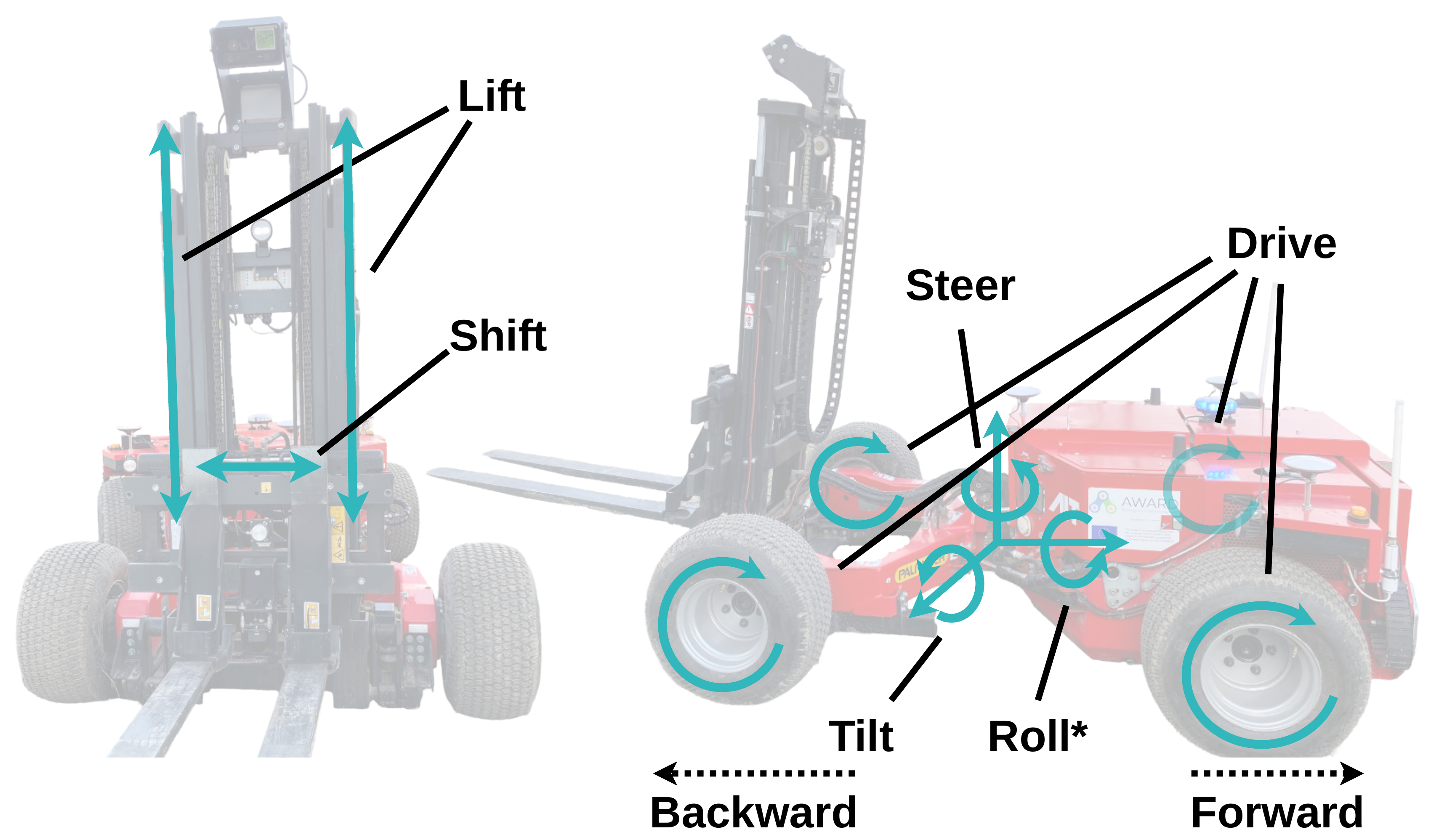}    
    \caption{Actuated and unactuated(*) joints for vehicle base movement and fork positioning.}
    \label{fig:actuators_joints}
\end{figure}

ADAPT is built on the Palfinger BM154 truck-mounted platform. 
Unlike common warehouse forklifts, this platform is remotely controlled, which eliminates the need for an operator seat. Its foldable design allows it to be stored beneath a truck's cargo area, making it an ideal solution for delivering goods to environments without established infrastructure. 
The vehicle is widely used in challenging operational settings in multiple industries, including the delivery of construction material to the worksites, the transportation of agricultural equipment and landscaping supplies, the logistics of beverages, and civil protection and disaster management.

Table~\ref{tab:bm154} lists the main technical specifications of the BM154.
The platform's primary strengths lie in its advanced all-terrain capabilities, enabled by an all-wheel drive system and its center-articulated steering with 3 rotational degrees of freedom (see Figure~\ref{fig:system_overview}a). These features allow for navigating in uneven terrain under various ground conditions such as asphalt, gravel, mud, or grass. 
The platform consists of two chassis parts, the front part housing the motor and the hydraulic block, and the rear part incorporating the forks attached to the lifting mast. 
The motor side is designated as the main driving direction for autonomous operation (see Figure~\ref{fig:actuators_joints}).
This is motivated by considerations of load handling safety and operational visibility due to the unobscured view.

ADAPT has been extensively customized for automation since 2019 as part of several funded national and international research projects. Figure~\ref{fig:system_overview} illustrates the current hardware setup, highlighting the added components required for autonomous operation. The following sections provide a detailed overview of the hardware architecture and emphasize the differences between the autonomous prototype and the platform for manual operation.

\begin{table}[h!]%
\caption{Specifications of the Palfinger BM154 platform.}
\label{tab:bm154}
\begin{tabular}{l | l}
\textbf{Safe Working Load}       & 1500kg @ 0.6m  \\ \hline
\textbf{Lift Height}             & 2.85m \\  \hline
\textbf{Weight}                  & Approx. 1700kg  \\ \hline
\textbf{Engine}                  & 3-cylinder Diesel, 18.8kW \\  \hline
\textbf{Drive System}            & Hydrostatic 4-wheel drive
\end{tabular}
\end{table}

\subsection{Actuation}
\label{sec:actuation}

As shown in Table~\ref{tab:bm154}, the BM154 platform uses a diesel engine as its main power source, driving two distinct hydraulic circuits that control vehicle movement and operational functions. 
The diesel engine drives a closed hydraulic circuit designed for operating the wheels and an open hydraulic circuit responsible for actuating multiple hydraulic cylinders, amongst others, for steering and lifting the forks. 
Figure~\ref{fig:actuators_joints} and Table~\ref{tab:interfaces} provide graphical and concise textual information on the forklift's actuated and non-actuated joints, driven by the hydraulic motor and cylinders.

The closed hydraulic circuit drives the vehicle's wheels, delivering continuous traction for efficient power transfer and enhanced mobility. 
This design optimizes vehicle stability and maneuverability, particularly in challenging or uneven terrains. 
The hydrostatic four-wheel drive system is controlled by a PWM signal, where the duty cycle determines the flow rate of the hydraulic fluid, which in turn directly governs the wheel speed.

The open system discharges hydraulic fluid after use into the tank, making it more suitable for functions that require intermittent actuation.
The open circuit governs several essential functions, including steering and tilting of the central joint that connects the front and rear chassis (see Figure~\ref{fig:system_overview}a). 
Although the rolling rotation of the chassis is passive and not directly actuated, it is fundamental to maintaining ground contact and load distribution on irregular terrain. 
The open circuit also facilitates mast actuation, allowing vertical fork movement (lifting and lowering) and lateral adjustments (shifting left and right). 
The control of the open hydraulic system's cylinders is facilitated by proportional valves, which allow precise control of valve spool positions via a CAN interface.

\begin{table}[h]%
    \caption{Interfaces for actuation and proprioceptive feedback for joint position, velocity, and hydraulic pressure. \tcmark ~indicates directly available interfaces, while \rxmark~denotes unavailable ones. \gcmark ~signifies feedback that is obtained through post-processing of sensor data.}
    \label{tab:interfaces}
    \centering
    \setlength{\tabcolsep}{8pt} 
    \begin{tabular}{l|c|c|c|c}
        & Active & Position & Velocity & Pressure \\
        & Actuation & Feedback & Feedback & Feedback \\
        \hline
        \textbf{Drive}  & \tcmark & \rxmark & \tcmark & \tcmark \\
        \hline
        \textbf{Steer}  & \tcmark & \tcmark & \gcmark & \tcmark \\
        \hline
        \textbf{Tilt}   & \tcmark & \tcmark & \gcmark & \tcmark \\
        \hline
        \textbf{Roll}   & \rxmark & \tcmark & \gcmark & \rxmark \\
        \hline        
        \textbf{Lift}   & \tcmark & \tcmark & \gcmark & \tcmark \\
        \hline
        \textbf{Shift}  & \tcmark & \tcmark & \gcmark & \tcmark \\
    \end{tabular}    
\end{table}

\subsection{Sensing}
\label{sec:sensing}
This section describes the proprioceptive and exteroceptive sensing hardware components of ADAPT.
Proprioceptive sensors monitor the vehicle's internal state, enabling precise joint control, while exteroceptive sensors facilitate accurate pallet and loading platform pose estimation, vehicle localization and mapping, and obstacle detection. 
Advanced algorithms that process sensor data and enable higher-level decision-making and motion control are discussed in Sections~\ref{sec:perception} and \ref{sec:planning_control}.

\subsubsection{Proprioception}
\label{sec:propioception}

The forklift's proprioception system integrates several sensors to monitor its internal state with high precision, especially the joint values depicted in Figure~\ref{fig:actuators_joints}. 
In terms of vehicle motion feedback, wheel speed encoders are used to monitor the rotational velocity of each wheel. 
A system utilizing three linear potentiometers captures the 3D rotation of the central articulated joint.
By measuring the sensor elongation, the three articulation angles can be calculated with sub-degree precision. 
The resulting values are integral for steering the vehicle and precisely adjusting the fork tilt angle during load manipulation.
Pressure sensors are installed within the hydraulic cylinders and drive system to measure pressure levels at high frequencies.
These measurements are crucial for controlling the forklift's load-handling capabilities, including detecting whether the fork is in direct contact with the environment.
In addition, draw-wire encoders are implemented to measure the vertical position of the lifting mast and the lateral movement of the forks (side-shift) with millimeter precision. 
Together, these sensors provide comprehensive real-time feedback and allow for accurate modeling of the behavior of the system.

\subsubsection{Exteroception}
\label{sec:exteroception}

In robotics, the exteroceptive system refers to the sensors and mechanisms that allow a robot to perceive and interpret information from its environment. For \ADAPT, the exteroceptive system is crucial for localization and mapping, precise object detection, traversability assessment, and obstacle avoidance, ensuring safe and efficient operations in complex environments. 

\begin{figure}[h]
    \centering
    \includegraphics[width=1\linewidth]{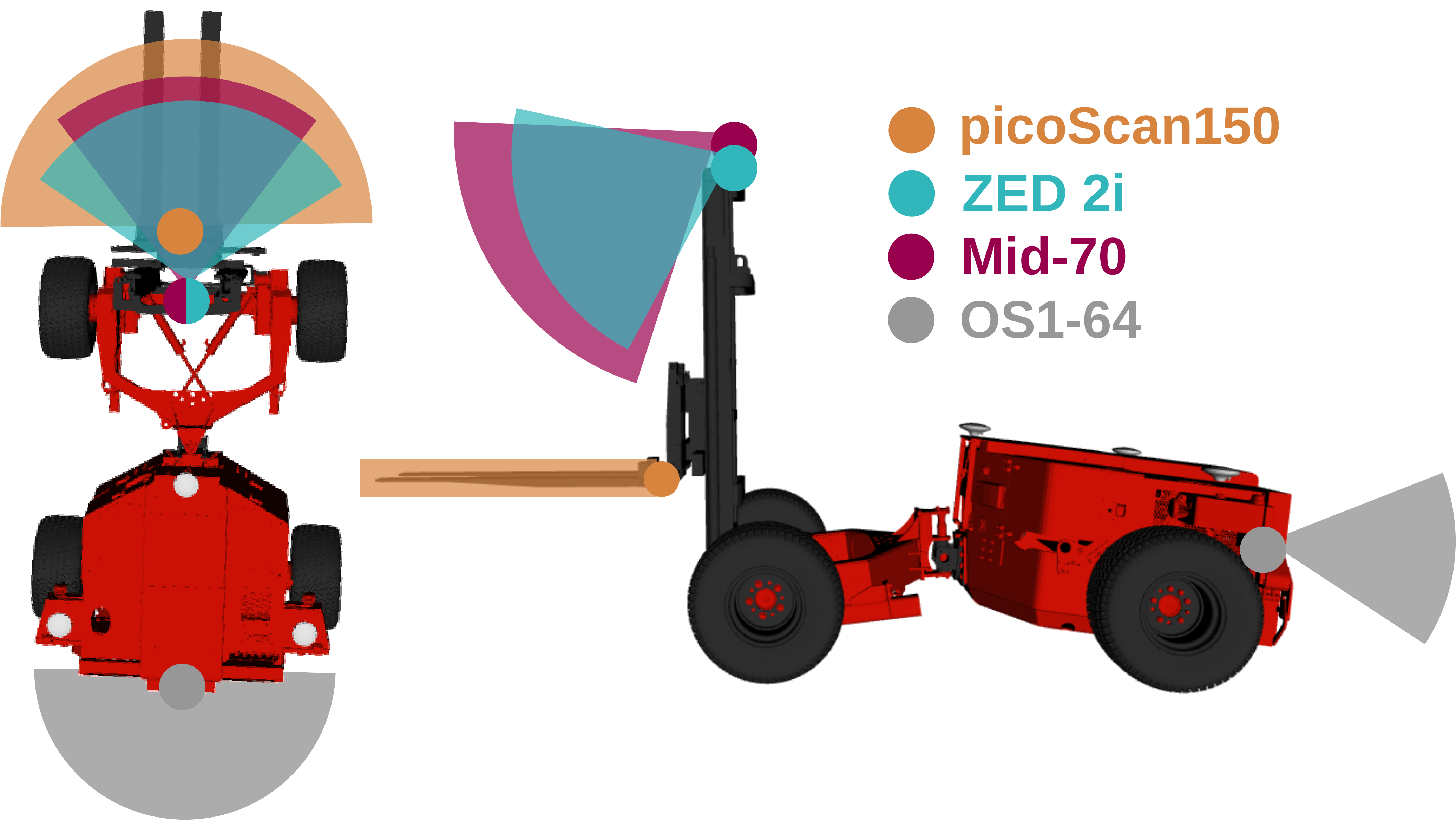}
    \caption{Placement of exteroceptive sensors and their respective fields of view.}
    \label{fig:sensor_fov}
\end{figure}

The placement of sensors on the machine is determined by the function of each sensor and the operational phase in which it is required, as shown in Figure~\ref{fig:system_overview}, \new{with the corresponding sensor fields of view illustrated in Figure~\ref{fig:sensor_fov}}. High-resolution 3D mapping of the environment and obstacle detection is performed by the wide field of view Ouster OS1-64 LiDAR, mounted in the machine's forward direction (see Figure~\ref{fig:system_overview}c). In fork/backward direction, a narrow field of view Livox Mid-70 LiDAR sensor, and a ZED2i stereo camera are employed for object detection (see Figure~\ref{fig:system_overview}b).
In the forward direction, the system moves faster and covers longer distances, increasing the risk of collisions with dynamic obstacles. Thus, active low-latency obstacle detection is crucial during path execution. In the backward direction, where pallets are only loaded and unloaded, explicit obstacle detection (going beyond terrain mapping) is not essential.
For pallet recognition, the ZED2i is the central component. It provides the input data for the neural networks for pallet recognition and pose estimation.
Additionally, for precise short-range detection during the final pallet approach, the forklift is equipped with a Sick picoScan150 2D LiDAR (see Figure~\ref{fig:system_overview}d). It ensures accurate positioning of the forks relative to the pallet, due to its high frequent update rate and well-aligned viewing angle.
The forklift utilizes the Livox LiDAR to enable accurate detection of the truck’s loading edge.
Additionally, a Septentrio mosaic-H system with two antennas is used for RTK GNSS localization. It provides accurate position information at the centimeter level and heading information to ensure precise alignment during loading and unloading tasks.

By integrating these sensors, the exteroceptive system equips \ADAPT with comprehensive situational awareness, supporting centimeter-accurate load handling and safe navigation in outdoor settings.

\subsection{Processing Components and Network}

\begin{figure} [h]
    \centering
    \includegraphics[width=1\linewidth]{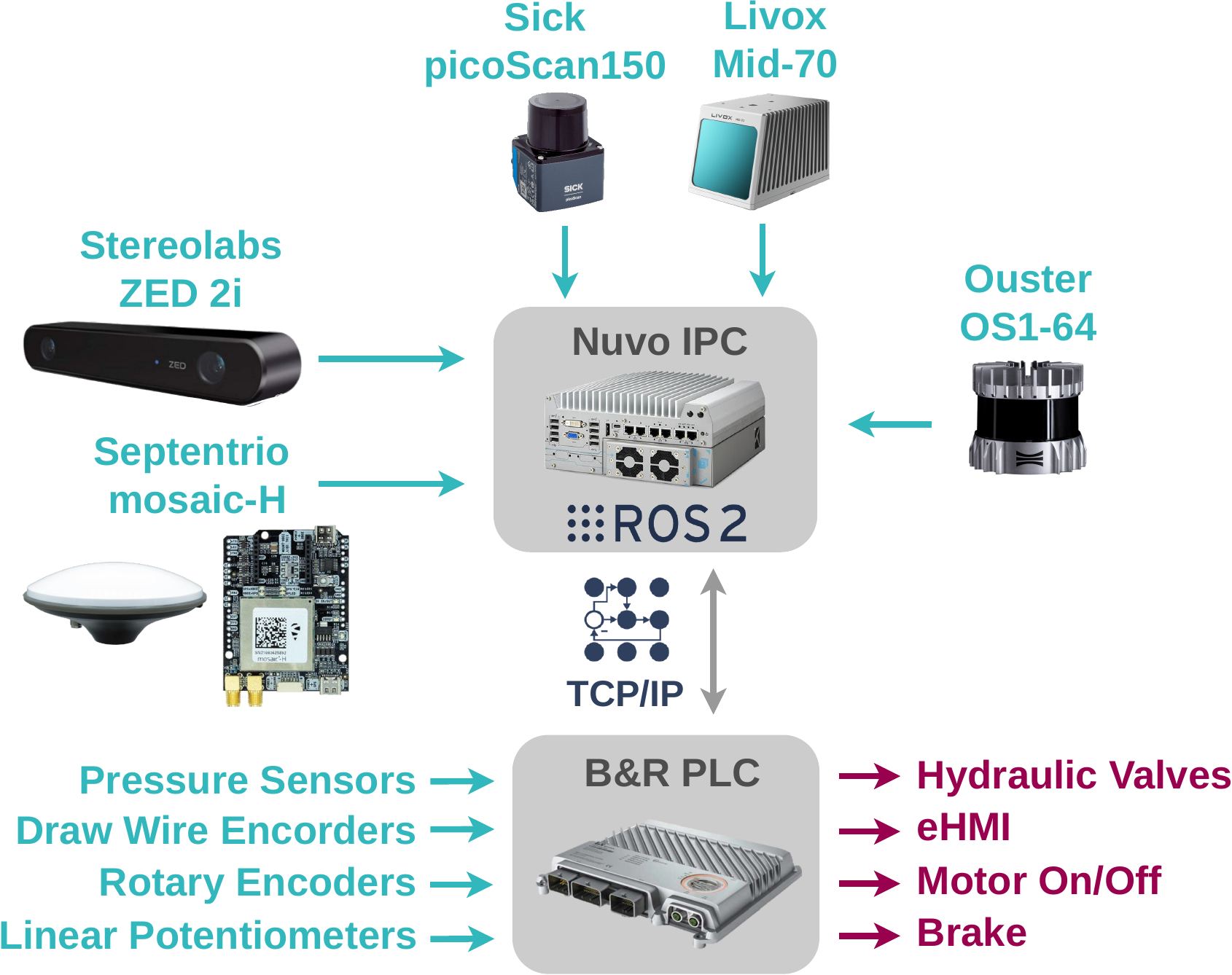}
    \caption{Overview of the main hardware components, including core processing devices (grey), actuators (red), as well as proprioceptive and exteroceptive sensors (cyan).}
    \label{fig:automation_components}
\end{figure}

\ADAPT is designed with a robust and distributed processing hardware architecture, depicted in Figure~\ref{fig:automation_components}.
The core system consists of two primary computational components: a Programmable Logic Controller (PLC), and a rugged industrial PC (IPC), each assigned distinct responsibilities.

The B\&R X90 PLC manages low-level hardware interfaces, including direct connections to sensors and actuators that require low latency and real-time control. It also handles safety-critical operations that depend on rapid response times, such as emergency stops.
The rugged IP67-compliant design of the PLC makes it suitable for harsh outdoor environments and does not require an additional enclosure.
For higher-level processing, such as planning, advanced control algorithms, and managing intelligent sensors, a Nuvo-9166GC IPC is employed.
It connects to the PLC via a central Ethernet switch to receive the low-level sensor data and to transmit high-level commands over a custom TCP/IP protocol.
This setup enables seamless integration and coordination between low-level operations and advanced computational tasks that do not require millisecond-level latency.
These advanced computational tasks include high-level task and motion control, vehicle state estimation, environment mapping, obstacle avoidance and object detection pipelines, utilizing camera and LiDAR sensors.

A wireless network bridge connects the forklift to a remote monitoring station housed in a weatherproof container, primarily supporting development and expert interventions. 
This setup enables live monitoring of the system's operation, ensuring efficient debugging and adjustments during testing.
Additionally, a supplementary WLAN antenna allows on-site operators to monitor the system state via a tablet-based Human-Machine Interface (HMI). 
For long-range communication, an LTE/5G router is integrated.
All components discussed, along with the necessary power supply electronics, are IP67 rated or housed within weatherproof switching cabinets, ensuring reliable operation under all typical weather conditions in Central Europe.

\begin{figure} [ht]
    \centering
    \includegraphics[width=1\linewidth]{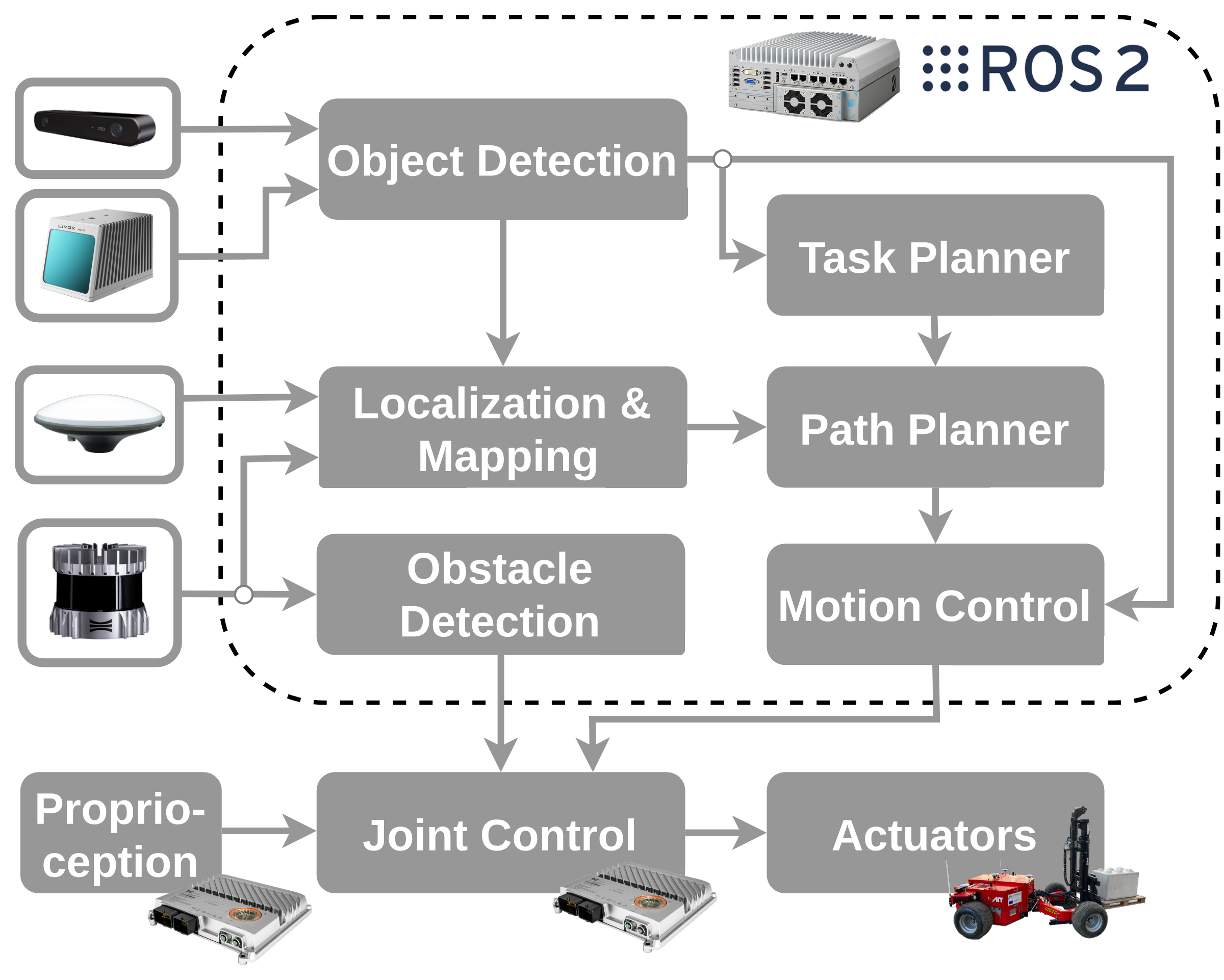}
    \caption{Overview of the main software components, with indicated data flow and respective processing devices.}
    \label{fig:software_overview_detailed}
\end{figure}

\subsection{Software Components}
Figure~\ref{fig:software_overview_detailed} shows an overview of the software architecture as well as concise descriptions of core functionalities and their associated hardware components.
Core functionalities include localization and mapping, object and obstacle detection, task and path planning, and motion and joint control.
Localization and mapping are based, amongst others, on GNSS modules and LiDAR to provide accurate environmental maps and vehicle positioning data.
Object detection utilizes a stereo camera and LiDAR to identify pallets and truck loading edges, facilitating precise load carrier handling. 
The obstacle detection module ensures safety by halting the vehicle before potential collisions with detected obstacles.
The task planner serves as the central planning component of the system, managing high-level decision-making and coordinating both the path planner and motion controller.
The path planner creates optimal routes based on map and localization data, while the cascaded motion controller provides precise vehicle base movements and responsive fork actuation for effective pallet manipulation.
The software components running on the rugged IPC highly utilize the Robot Operating System (ROS 2) \cite{ROS2_2022} middleware and its associated libraries, which enable a decentralized architecture while ensuring robust data and information exchange.
A detailed discussion of the modules presented can be found in Sections \ref{sec:perception} and \ref{sec:planning_control}.

\section{Perception}
\label{sec:perception}

\ADAPT relies on multiple perception modules to function effectively in an outdoor environment. 
The first requirement is self-localization within its surroundings, which is integrated with mapping of pallet poses in a simultaneous localization and pallet mapping process, as detailed in Section~\ref{sec:state_estimation}.
However, a more detailed environment mapping is necessary to assess terrain traversability, as shown in Section~\ref{sec:mapping}. 

Since autonomous exploration on a public construction site is infeasible, we assume an initial human-guided exploration phase. 
A human controller then designates general loading and unloading areas to define the mission: which pallets should be brought where, and how many. Within these areas, ADAPT autonomously detects all pallets and trucks and estimates their 6D poses. The pallet detection process is described in Section~\ref{sec:pallet_detection}, while the estimation of the edge of loading of the truck is covered in Section~\ref{sec:loading_edge_detection}.
Lastly, to ensure safe navigation and prevent collisions during path execution, the forklift must detect and take into account obstacles in its surroundings, as discussed in Section~\ref{sec:obstacle_detection}.


\subsection{Joint Localization and Pallet Mapping}
\label{sec:state_estimation}
 The goal of mapping is to derive a representation of the environment that is as rich in features as necessary for the application, but as compact as possible to keep computational complexity low. 
For ADAPT, there are two levels of abstraction in which we need to map the environment: The high-level representation of the pallet poses and the lower-level environment map with a strong focus on terrain traversability assessment, as described in Section~\ref{sec:mapping}.

\begin{figure}[t]
    \centering
\begin{tikzpicture}

\tikzstyle{gnss}=[draw=blue,fill=blue!20]
\tikzstyle{odom}=[draw=red,fill=red!20]
\tikzstyle{pallet}=[draw=green,fill=green!20]

\node[latent] (X0) at (0,0) {$X_0$};
\node[latent] (X1) at (1.5,0) {$X_1$};
\node[latent] (X2) at (3,0) {$X_2$};
\node[latent] (X3) at (4.5,0) {$X_3$};
\node[latent] (X4) at (6,0) {$X_4$};

\factor [odom, right=0.25 of X0] {oO} {$O_{0,1}$} {X0} {X1};
\factor [odom, right=0.25 of X1] {o1} {$O_{1,2}$} {X1} {X2};
\factor [odom, right=0.25 of X2] {o2} {} {X2} {X3};
\factor [odom, right=0.25 of X3] {o3} {} {X3} {X4};

\factor[gnss, above=0.5 of X0] {G0} {$G_0$} {} {X0}
\factor[gnss,above=0.5 of X2] {G2} {$G_2$} {} {X2}

\node[latent, pallet] (P0) at (1.5,-1.5) {$P_0$};

\factor[pallet, above=0.4 of P0] {G0} {} {X1} {P0}
\factor[pallet, below=0.6 of o1] {G0} {} {X2} {P0}

\node (lp1) at (3.5,-1.7) {};
\node (lp2) at (5,-1.7) {};

\node (lg) at (6.5,-1.7) {};
\node (lo1) at (5,-1) {};
\node (lo2) at (6.5,-1) {};
\factor[pallet, right=0.5 of lp1] {G0} {Pallet obs.} {lp1} {lp2}
\factor[gnss, left=0.55 of lg] {G0} {GNSS} {} {lg}
\factor[odom, right=0.5 of lo1] {G0} {Odometry} {lo1} {lo2}
\end{tikzpicture}
    \caption{Factor Graph for the joint forklift poses ($X_x$) and pallet poses ($P_p$) over time (left to right), relative (from one time step to the next) odometry information $O_{p1,p2}$, and GNSS measurements $G_g$. In this example, the pallet $P_0$ was observed in timesteps $1$ and $2$, and GNSS information was available in timesteps 0 and 2.}
    \label{fig:graph_state_estimation} 
\end{figure}
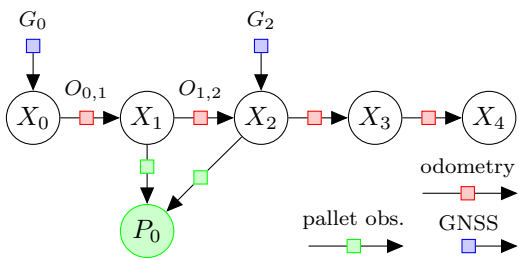

We use a factor graph-based approach~\cite{kschischang2001factor} that integrates the vehicle’s odometry data, GNSS measurements, and pallet detections into a joint optimization framework for high-level localization and pallet mapping.
A factor graph is a theoretical framework that models unknown variables, measurements, and their interdependencies as a bipartite graph with two types of nodes (see Figure~\ref{fig:graph_state_estimation}). 
It consists of variable nodes (depicted as circles), which represent the variables to be estimated, and factor nodes (depicted as rectangles). In this context, factors can be binary (functions that relate two unknowns with one another) or unary (constraints on a variable based on a measurement).
This factor graph approach is advantageous in this context as it allows flexible integration of multiple types of measurement and provides a probabilistic framework for managing uncertainties inherent in noisy and delayed sensor data. 
Our factor graph is shown in Figure~\ref{fig:graph_state_estimation}.

Under outdoor conditions, dual-antenna RTK-GNSS provides global position and attitude, which are transformed into UTM coordinates and referenced to a local site origin, with receiver-reported Gaussian covariance used directly. Forklift pose nodes are connected by odometry derived from wheel encoder and steering angle measurements, which are modeled as zero-mean Gaussian noise with variances of $0.1 m/s$ in translational velocity and $0.5^{\circ}/s$ in steering angular velocity, to account for slip and kinematic errors. 
Pallet detections are incorporated as binary pose factors between the pallet and forklift poses, with a diagonal Gaussian covariance. The position components are expressed in the camera coordinate system, with $0.05m$ uncertainty in the lateral directions (parallel to the image plane) and $0.10m$ in the depth direction (along the viewing axis). The orientation components are expressed in the pallet coordinate system, with $2^{\circ}$ uncertainty in roll and yaw, which are well-constrained by the pallet front edge, and $5^{\circ}$ in pitch, which is less constrained due to the potential occlusion of the pallet’s back corners. To account for degradation with viewing distance, the position and orientation uncertainties increase linearly with range at rates of $0.02m/m$ and $0.5^{\circ}/m$, respectively.

We build upon iSAM2~\cite{kaess2012isam2} incremental smoothing to maintain real-time performance, yielding the most probable pallet and forklift poses given all odometry, GNSS, and pallet measurements.
This allows the pose of a pallet to be optimized by measurements from different views. In addition, the pallet poses and vehicle location are optimized even without or with a less precise GNSS signal. GNSS measurements are discarded if the reported variance exceeds $0.1m$ or if a position jump greater than $0.5m$ is detected. In typical deployment environments, signal quality is presumably good, and prolonged GNSS outages are neither expected nor permitted in order to remain within the GNSS-defined mission zones.

With iSAM2 updates of pallet detections and localization information are optimized incrementally, in contrast to repeatedly resolving the entire optimization. To do so, iSAM2 maintains a dynamic Bayesian tree structure to track variable interdependencies in addition to the factor graph.

This approach ensures that updates to the factor graph are confined to regions where interdependency changes occur. To facilitate real-time operation, our iSAM2-based optimization approach for joint pallet and vehicle pose mapping
\begin{compactitem}
    \item linearizes the problem around an initial guess,
    \item reinitializes (recomputes the linearization) when a variable changes significantly,
    \item exploits the sparsity of the optimization problem, since each factor connects only a small subset of variables, leading to a sparse Jacobian matrix.
\end{compactitem}

Beyond iSAM2, our approach applies pose pruning and culling for measurements older than a user-specified maximum sensor latency ($5s$ in our setup), adopting a fixed-lag smoothing strategy to bound the measurement history and maintain computational efficiency. Culling operates on a pose similarity threshold, merging forklift poses with translational difference below $1m$ and rotational difference below $5.7^{\circ}$ into a single forklift node. Nodes with associated pallet detections are preserved. The $5s$ window is selected to guarantee temporal alignment of all sensor measurements, including those from high-latency detectors, before pruning is applied.

This approach can efficiently integrate information from odometry sensors, GNSS, and detections over time to yield, given all sensor information, the most probable result for the unknowns (the maximum a posteriori estimate). The pallet mapping is tightly integrated into our optimization framework. For pallet tracking, we use a Mahalanobis distance-based~\cite{ReprintMahalanobisPC2018} approach to ensure a robust association of detections between consecutive measurements. The Mahalanobis distance metric allows for 
dynamic association by accounting for sensor pose uncertainty, as well as uncertainty in pallet  pose estimation.
Additionally, pallets are dynamically managed within the forklift's view frustum by pallet bookkeeping that goes beyond pure variable marginalization (pruning). It manages the tracking of new never-before-seen pallets, associates new detections with old ones, and deletes pallets if they should be in sensor view but cannot be detected for a while.
This approach aligns with the concept of object permanence, where pallets are presumed to remain at a designated location even when they are not directly visible.

\new{
\textbf{Evaluation of Joint Localization and Pallet Mapping:} 
To assess the extent to which pallet observations can compensate for the absence of GNSS updates, we evaluate the approach on 10 independent recordings of the forklift operating near pallets at the test site introduced in Section~\ref{sec:experiments}. During these recordings, no pallet is contacted, picked up, or moved; hence, manually surveyed pallet poses remain valid as static ground truth throughout each run. The recordings contain representative maneuvers near a pallet, including approach, departure, short stops, and reversing. They span durations from $28\mathrm{s}$ to $100\mathrm{s}$ and traveled distances from $14\mathrm{m}$ to $39\mathrm{m}$.
Each recording is replayed offline under three estimator configurations, while keeping all estimator parameters and pallet-bookkeeping rules unchanged:
\begin{compactitem}
\item \textit{odom}: wheel and steering odometry only.
\item \textit{odom+pallets}: odometry and online pallet detections fused in the factor graph.
\item \textit{full}: GNSS, odometry and online pallet detections.
\end{compactitem}
For \textit{odom} and \textit{odom+pallets}, GNSS is used only during the first $2\mathrm{s}$ to initialize the graph in the UTM frame and is then disabled. This emulates operation after GNSS loss following initial localization.
}

\begin{table}[ht]
\centering
\caption{\new{Forklift-to-pallet position error after GNSS initialization (RMSE). \textit{odom+pallets} uses online pallet detections; \textit{odom+surveyed} uses the odometry-only vehicle trajectory and the surveyed pallet position, without online pallet updates. $N$ is the number of stereo pallet detections of the target pallet admitted to the graph after the GNSS initialization burst. $\Delta$ is the relative change of \textit{odom+pallets} with respect to \textit{odom+surveyed}}.}
\label{tab:runs_pick}
\setlength{\tabcolsep}{3.5pt}
\small
\begin{tabular}{lrrrrrr}
\toprule
Run & Len. & Dur. & $N$ & \textit{odom+pallets} & \textit{odom+surveyed} & $\Delta$ \\
    & [m]  & [s]  &     & [m] & [m] & [\%] \\
\midrule
S1  & 23 & 100 & 42 & 0.054 & 0.118 & -54 \\
S2  & 19 &  84 & 32 & 0.039 & 0.138 & -72 \\
S3  & 19 &  49 & 24 & 0.069 & 0.175 & -61 \\
S4  & 20 &  66 & 34 & 0.075 & 0.146 & -49 \\
S5  & 23 &  28 & 11 & 0.055 & 0.158 & -65 \\
S6  & 39 &  48 & 14 & 0.052 & 0.289 & -82 \\
S7  & 31 &  39 & 10 & 0.054 & 0.458 & -88 \\
S8  & 21 &  32 & 11 & 0.043 & 0.100 & -57 \\
S9  & 20 &  46 & 19 & 0.090 & 0.128 & -30 \\
S10 & 14 &  39 & 22 & 0.105 & 0.111 &  -6 \\
\midrule
Mean & 23 &  53 & 22 & 0.064 & 0.182 & -65 \\
\bottomrule
\end{tabular}
\end{table}

\new{
\textbf{Localization accuracy:}
The absolute translational trajectory error (ATE-trans, RMSE) is first computed against the RTK-GNSS vehicle trajectory. Since the \textit{full} configuration fuses the same RTK-GNSS stream that serves as the trajectory reference, all \textit{full}-configuration results are interpreted as consistency checks rather than as independent validation. The two GNSS-denied configurations provide the relevant localization ablation. Pure odometry exhibits increasing drift with traveled distance, reaching $0.37\,\mathrm{m}$ on the longest run (S6, $39\,\mathrm{m}$), corresponding to approximately $1\%$ of the path length. Adding pallet detections leaves global ATE at a comparable level, as expected: a pallet landmark initialized from a drifted vehicle pose constrains the relative vehicle-pallet geometry, but cannot by itself remove accumulated global drift.
}

\new{
\textbf{Pallet-relative accuracy:}
For pallet handling, the operationally relevant quantity is not the global vehicle pose but the relative pose between the forklift and the target pallet. The estimated forklift-to-pallet position after GNSS initialization is therefore compared against a reference obtained from the RTK-GNSS vehicle trajectory and the surveyed pallet pose. Table~\ref{tab:runs_pick} compares \textit{odom+pallets} with the \textit{odom+surveyed} baseline (defined in the caption). Across all 10 runs, online pallet detections reduce the mean forklift-to-pallet position RMSE from $0.182\,\mathrm{m}$ to $0.064\,\mathrm{m}$, indicating a mean reduction of $65\%$, with the smallest improvement on the shortest run (S10). Pallet observations thus primarily improve the manipulation-relevant relative estimate rather than the globally referenced trajectory.
}

\new{
\textbf{Pallet-map consistency:}
As a compact map-quality check, the optimized pallet position in the \textit{full} configuration is compared against the surveyed position. The resulting XY error ranges from $0.014\mathrm{m}$ to $0.084\mathrm{m}$ (mean $0.039\mathrm{m}$). As above, this is a consistency check since the RTK-GNSS stream anchors the world frame. Nevertheless, the optimized landmarks converge to stable positions consistent with the manual survey, and no duplicate track for the same physical pallet occurred in any run. Overall, the ablation confirms the intended role of the pallet factors: they do not remove the global drift that accumulates after GNSS loss, but they substantially improve the relative forklift-to-pallet estimate required for pallet handling.
}

\subsection{Traversability Mapping}
\label{sec:mapping}

\begin{figure*}
    \centering
    \includegraphics[width=1\linewidth]{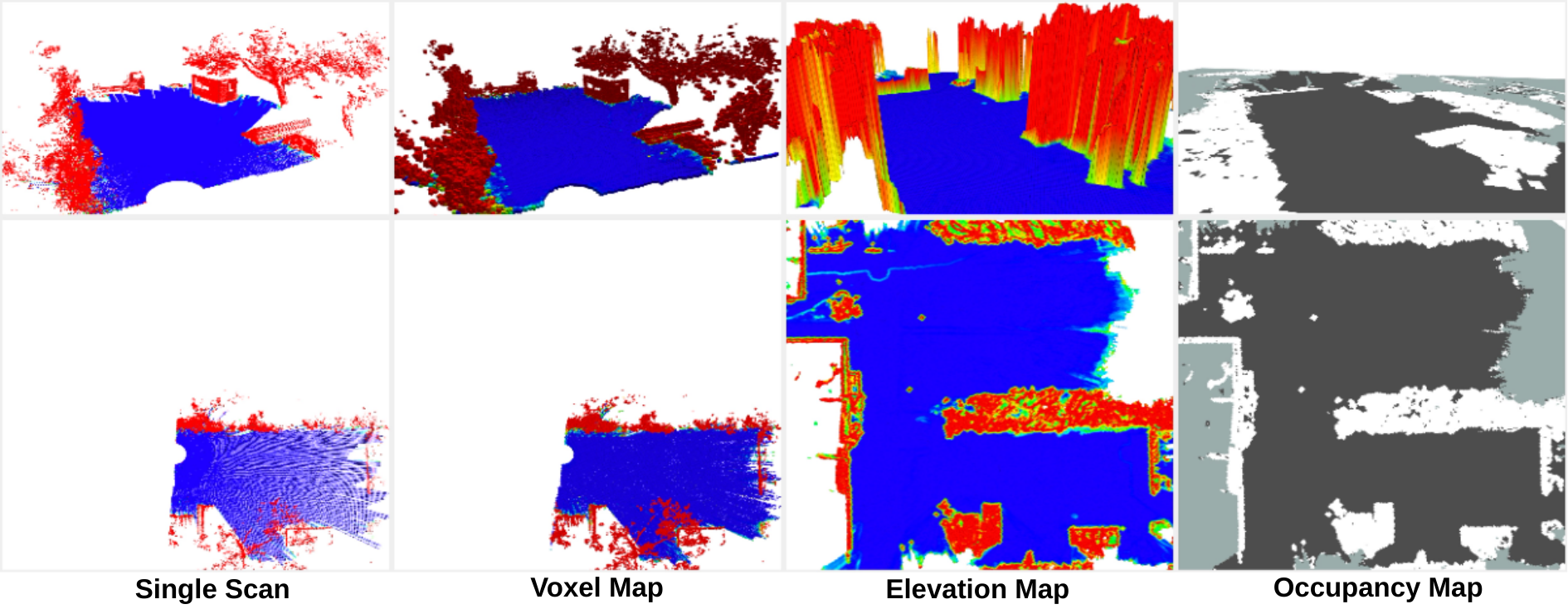}
    \caption{Mapping stages. Top row: 3D view. Bottom row: Top view. The color shows the traversability from blue (good) to red (bad).}
    \label{fig:mapping}
\end{figure*}

The targeted areas of application, e.g., construction sites, of \ADAPT are unstructured environments with potentially uneven terrain. Hence, a specific environment mapping targeting these aspects is required.
Therefore, a 2.5D elevation map representation is generated from the 3D point measurements of the Ouster LiDAR. While path planning complexity can be reduced to planning in 2 dimensions, the most relevant geometric information about the environment must still be available from the map. 
In particular, it must allow for a clear distinction between traversable areas, obstacles, and slopes.
This distinction is crucial for forklifts, as they are more susceptible to tipping over when carrying loads.

Similarly to \textit{spatio-temporal voxel layers}~\cite{macenski_spatio-temporal_2020}, our approach aggregates data into a 3D voxel grid with short-term memory. However, instead of directly deriving a 2D cost map, our method employs a 2.5D elevation map for dynamic obstacle handling and terrain analysis, serving as a long-term memory representation.
Therefore, it is implemented as a three-stage process, encompassing the analysis of single scans, 3D voxel map aggregation, and the generation of a 2.5D elevation map. In all three steps, distinguishing between the classes ground and obstacle plays a decisive role.

\textbf{Analysis of single scans:} 
Single scans are analyzed within the sensor domain, utilizing the unique capabilities and characteristics of the sensor. By focusing on individual scans, this stage minimizes disturbances caused by alignment errors. The probability $P(\mathbf{x}\mid\cdot)$ of a point $\mathbf{x}$ to be an obstacle 
\begin{equation}
P(\mathbf{x}\,|\,h, \alpha) := P(\operatorname{class}(\mathbf{x}) = \text{obstacle}\,|\,h, \alpha)
\end{equation}
is computed using the observations of the height 
\begin{equation}
h = \operatorname{dist}\left(\mathbf{x}, \text{ground plane}\right)
\end{equation}
of the point $\mathbf{x}$ above the ground plane and the slope angle
\begin{equation}
\alpha = \arccos\left({\frac{\mathbf{n} \cdot \mathbf{g}}{\lVert\mathbf{n}\rVert \lVert\mathbf{g}\rVert}}\right)
\end{equation}
between its surface normal vector $\mathbf{n}$ and the gravity vector $\mathbf{g}$.

The initial ground plane is determined from LiDAR-vehicle extrinsic calibration and initialized using the forklift's wheel contact points, obtained from proprioception. Points close to that plane are then used for plane fitting. The fitted plane serves as the reference ground plane for computing $h$.

Under the assumptions of a uniform prior on the class, $P(\operatorname{class}(\mathbf{x}) = \text{obstacle})=0.5$, and conditional independence of the two observations, using the log-odds representation
\begin{equation}
\operatorname{logodds}(p) = \log\left(\frac{p}{1-p}\right),
\end{equation}
the combined probability $P(\mathbf{x} | h, \alpha)$ can be expressed as the sum of the individual log-odds 
\begin{equation}
\operatorname{logodds}\left(P(\mathbf{x}\,|\,h, \alpha)\right) = \operatorname{logodds}\left(P(\mathbf{x}\,|\,h)\right) + \operatorname{logodds}\left(P(\mathbf{x}\,|\,\alpha)\right)
\end{equation} 
with the following sensor models $P(\mathbf{x}\,|\,h)$ and $P(\mathbf{x}\,|\,\alpha)$ given by
\begin{align}
P(\mathbf{x}\,|\,h) &= \frac{1}{2} \tilde{p}_{h} + \frac{1}{2} \tilde{p}_{h}^2 \\
P(\mathbf{x}\,|\,\alpha) &= \frac{1}{2} \tilde{p}_{\alpha} + \frac{1}{2} \tilde{p}_{\alpha}^2,
\end{align}
where $\tilde{p}_{h}$ and $\tilde{p}_{\alpha}$ denote normalized intermediate terms derived from height and angle measurements and are defined as 
\begin{align}
\tilde{p}_{h} &= \min\left(\max\left(\frac{h}{h_{\max}}, 0.1\right), 0.9\right) \\
\tilde{p}_{\alpha} &= \min\left(\max\left(\frac{\alpha - \alpha_{\min}}{\alpha_{\max}-\alpha_{\min}}, 0.1\right), 0.9\right).
\end{align}
Using both $\tilde p_h$ and $\tilde p_h^2$ (and analogously for $\alpha$) yields a smoother increase near $0$ and a steeper one close to $1$. To avoid numerical instability ($\pm\infty$) in the log-odds computation, the values are constrained to the interval $\left[0.1, 0.9\right]$. 
Points with $P(\mathbf{x}\,|\,h,\alpha)>0.5$ (equivalent to $\operatorname{logodds}(P(\mathbf{x}\,|\,h, \alpha))>0$) are classified as obstacles, otherwise as ground.
In the forklift configuration, $h_{\max}=0.4\,\text{m}$ and $[\alpha_{\min},\alpha_{\max}]=[5^{\circ},\,30^{\circ}]$ (used internally in radians).
These thresholds were determined empirically by sweeping candidate values on a calibration dataset collected in our test site and choosing those that allowed discrimination of crucial drivable areas versus obstacles while avoiding false positives from gentle ramps and slightly tilted floors.

For the Ouster LiDAR, a single swipe of the nearly 180° field of view constitutes a scan with an organized raster structure, which is used for defining local neighborhoods for plane fitting in order to estimate the surface normals. The ego-motion during this rotation is compensated by transforming all points into the sensor's coordinate system at a fixed timestamp.

\textbf{3D voxel map aggregation}:
The second stage involves short-term aggregation of single-scan point clouds into a 3D voxel map, which can be performed based on time intervals or distance traveled. This process improves point density and addresses gaps caused by the LiDAR's sampling patterns. Additionally, statistical methods are used to remove outliers. The number of independent LiDAR scans hitting a voxel is stored for each voxel. A voxel is discarded if it has been hit by less than $m$ out of the last $n$ scans. For the forklift, short-term aggregation is over at least $0.25s$ and $1m$ traveled distance, requiring each voxel to be hit by at least $2$ out of the last $4$ scans. 

Overhanging structures, which present challenges for elevation maps, are identified by analyzing the free space between the ground and these structures. The voxels are traversed from top to bottom in vertical direction. If the unoccupied space between voxels is larger than the height of the vehicle, the higher voxels are discarded. The resulting output is an update for the elevation map, where each cell in the horizontal plane contains a computed elevation value and obstacle probability derived from the vertical voxel stack.

\textbf{2.5D elevation map}: 
The 2.5D elevation map integrates updates from the voxel map to form a long-term representation of the environment. 
The log-odds values of the obstacle probabilities of the update points and corresponding map cells are summed, and the classes are compared, which may lead to confirmation of classes or discrepancies. If the class is confirmed, the update is merged with the map cell, and the obstacle probability is converging towards $0$ or $1$. In the case of discrepancies, the obstacle probability will approach $0.5$ or even pass it, and the classification of a cell can change from obstacle to ground, or vice versa. A class change is interpreted as a change in the scene, which invalidates the cell's history, and the affected cell is completely replaced with the new values. This approach allows the map to adapt to changes such as the appearance, disappearance, or movement of objects. Adaptability is crucial for robots, such as forklifts, that actively manipulate their environment. The final traversability analysis combines obstacle probabilities with slope information derived from the elevation map's geometry. A tiling scheme ensures that the system can function in unbounded environments. 

The final output of the mapping process is an occupancy map containing critical spatial information, which serves as a 2D input for collision-free path planning. 
Figure \ref{fig:mapping} shows example views of the mapping stages at our test site in the outdoor laboratory, where the evaluations in Section~\ref{sec:experiments} were performed.


\subsection{Pallet Detection and Pose Estimation}
\label{sec:pallet_detection}
\begin{figure}[b]
    \begin{subfigure}[t]{0.48\linewidth}
        \centering
        \includegraphics[width=0.8\linewidth]{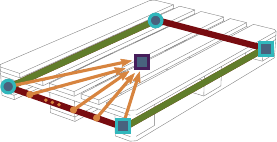}
        \caption{Part constellation definition.}
        \label{fig:pallet_ml}
    \end{subfigure}
    \hspace{0.02\linewidth}
    \begin{subfigure}[t]{0.48\linewidth}
        \centering        \includegraphics[width=0.8\linewidth]{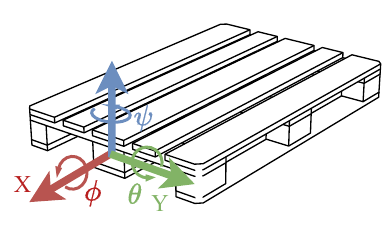}
        \caption{Definition for manipulation.}
        \label{fig:pallet_manip}
    \end{subfigure}
    \caption{Different pallet representations: For machine learning (a), the pallet is composed of two corner types (circle/rectangle) and two line classes (red/green) voting (orange) for a pallet center. In contrast, for manipulation (b), the origin of the pallet is located at the center of the approach side (see labels in Figure~\ref{fig:pallet_det}).}
    \label{fig:pallet}
\end{figure}

Detecting pallets and accurately determining their 6D pose enables \ADAPT to interact with them effectively. This is particularly important in construction sites, where pallets are typically placed arbitrarily rather than following a fixed grid or placement schedule.
The pallet detection methodology used is described in detail in \cite{beleznai_2024}. Given a dense stereo depth image as input, we intend to detect and estimate the 6D pose of multiple pallet instances with known structure and dimensions in a scene. To derive and exploit a learned neural representation, we proceed as follows:

\textbf{Pallet representation:} Pallets are represented by a part-constellation model (see Figure~\ref{fig:pallet_ml}) where pallet corners and edges not only form a localized part but at every pixel, they also define a spatial vote offset attribute, implying the offset vector to the corresponding object center. Part locations and local vote vectors are learned solely from synthetic data.

\begin{figure*}[h!]
    \centering
    \includegraphics[width=1.0\linewidth]{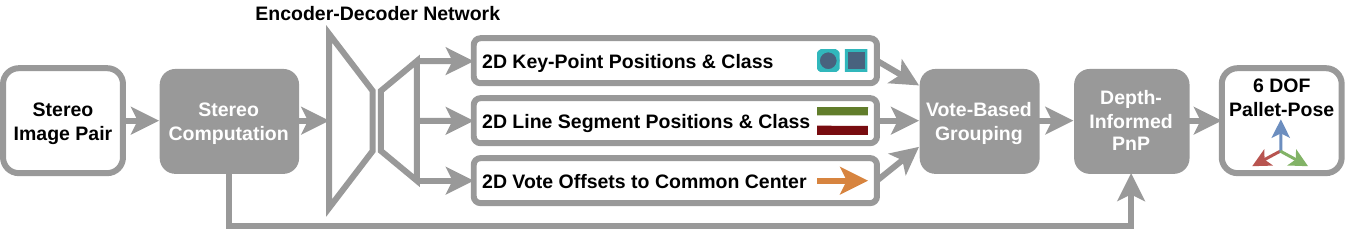}
    \caption{Overview of the learning/inference pipeline estimating multiple 2D parts used for 3D pose estimation (see symbols in Figure~\ref{fig:pallet}).}
    \label{fig:pallet_det}
\end{figure*}

\textbf{Data generation:} A synthetic data generation pipeline is used where more than 150 thousand stereo image pairs are generated with diverse pallet configurations, e.g., orientations, items carried, clutter. For details see \cite{beleznai_2024}. The data generation step relies on a simulated stereo camera setup, matching the baseline and optics of a ZED2i stereo camera~\cite{ZED2}. The synthetic stereo image pairs are used in a stereo depth computation step, resulting in depth data exhibiting similar characteristics to depth computed from real stereo pairs. This low sim-to-real gap allows for generating a wide variety of view variations and pallet configurations for learning. 

\textbf{Learning:} We employ an extensible convolutional encoder-decoder framework, based on~\cite{Zhou2019}, which estimates outputs for multiple learning tasks (see Figure~\ref{fig:pallet_det}). Pallets are represented by a part-based representation, where each part also carries a spatial vote offset attribute, where the offset vector points to the corresponding object center. During training, key-point and line segment representations yield multichannel heatmap output representations, where each channel is assigned to respective corner/edge categories.

\textbf{Detection and pose estimation during inference:} During inference, we use the stereo depth input of the ZED2i on-board camera to detect part-instances of pallet objects, where inferred parts are used to vote for pallet centers in a probabilistic manner, orange arrows in Figure~\ref{fig:pallet_ml}. 
Careful analysis of extensive key-point-based pose estimation experiments has revealed that detected points located on the approach side 
are more accurately determined than those farther away from the sensor. To mitigate the effect of spatially uneven distribution of re-projection errors on pose estimation, we set the pallet's origin to the center of the approach side, as illustrated in Figure~\ref{fig:pallet_manip}. For more details and pallet detection quality evaluations, see \cite{beleznai_2024}.

To estimate the 6D pose of the pallet, we formulate a nonlinear least-squares problem with the camera pose as the unknown and the pallet corners as fixed landmarks. Corner observations on the visible approach side are constrained by stereo camera projections, while occluded back-side corners are incorporated through monocular projections.
Since pallets may exhibit slight dimensional variations and corner detections are subject to uncertainty, the estimated camera-to-pallet distance can be biased. To address this, we apply a refinement step in which only the translation along the camera’s viewing direction is adjusted. This correction relies directly on a stereo depth measurement at the pallet origin (i.e., the front midpoint). By constraining the adjustment to the viewing direction, the accuracy and stability of the estimated pallet distance are improved while lateral and rotational components remain unaffected.

\begin{figure}[h]
    \centering
    \includegraphics[width=0.95\linewidth]{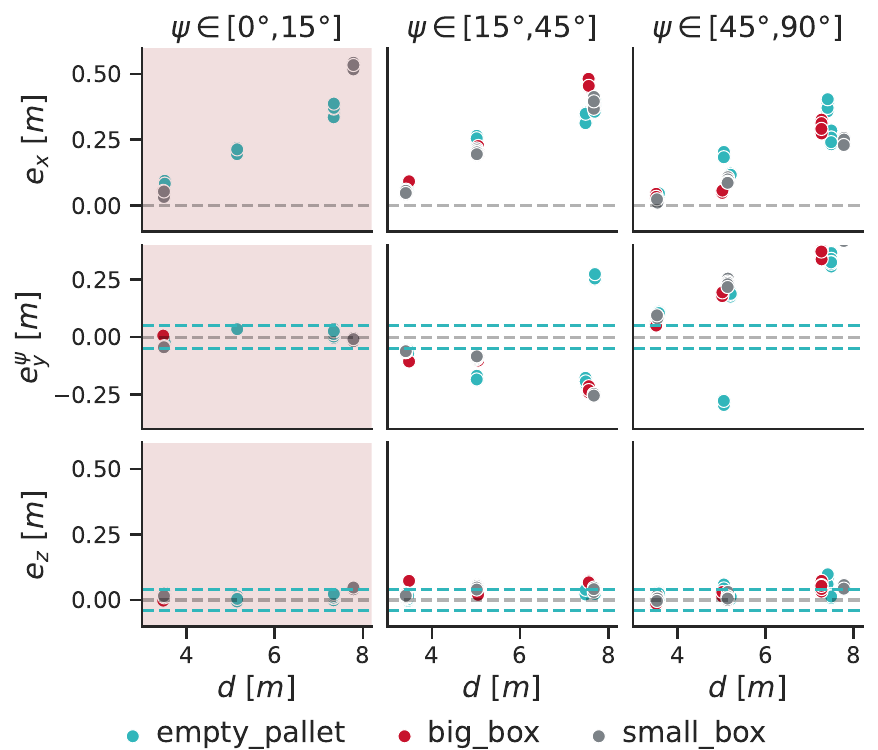}
    \caption{Pallet pose estimation errors over the distance to sensor $d$ for various load types. Columns represent different orientations around the pallet’s \textit{z}-axis. Dashed lines indicate tolerance boundaries for successful fork insertion, with the highlighted $\psi \in [0^\circ, 15^\circ]$ range being critical for insertion.}
    \label{fig:pallet_accuracy}
\end{figure}

\textbf{Pose estimation accuracy:}
By analyzing the geometry of Euro-pallets and fork dimensions, precise tolerance values for pallet insertion are established. Consequently, pose estimation errors serve as a key metric for evaluating the effectiveness of the proposed approach in achieving accurate insertion.
The ground truth data for error computation was obtained through manual annotation of a point cloud captured by an extrinsically calibrated LiDAR sensor mounted adjacent to the camera. Figure~\ref{fig:pallet_accuracy} presents the evaluation results based on 170 detections, covering various load types at different distances and orientations  around the pallet’s \textit{z}-axis.
To compare a detection with its corresponding annotation, the estimated 6D pose was transformed into a coordinate system centered at the manually annotated pose, cf. Figure~\ref{fig:pallet_manip}, yielding a 6D error vector $\mathbf{e}=[e_x,e_y,e_z,e_\phi,e_\theta,e_\psi]^\mathrm{T}$.
The tolerance limits for the relevant degrees of freedom are $y_{tol}=\pm0.05m$ and $z_{tol}=\pm0.04m$, with respect to the pallet coordinate system.
The constraint in the \textit{x}-direction is theoretically determined by the load’s center of mass, however, complete fork insertion is preferable to ensure optimal transport stability.
Assuming the pallets remain parallel to the ground, detections with unsuitable roll and pitch angles are filtered out, as the forklift cannot successfully pick them up.
Thus, these dimensions are excluded from the evaluation.
Errors in the y-direction \( e_y \) are affected by deviations in orientation around the pallet's \textit{z}-axis, which modify the effective pallet opening by a factor of $\cos(e_\psi)$. To account for this influence, the corrected \textit{y}-direction error is computed as $e_y^{\psi}=\frac{e_y}{\cos(e_\psi)}$.

The results indicate that detection accuracy deteriorates with increasing sensor distance, particularly in \( e_x \), due to the quadratic increase in depth error inherent in stereo vision. Similar degradation can be observed for recall values on a keypoint level (\cite{beleznai_2024}, Figure 8).
Furthermore, when the pallet is rotated around its \textit{z}-axis, the error is distributed between \( e_x \) and \( e_y \).
However, since the fork direction is aligned with the sensor's viewing direction around the pallet's \textit{z}-axis, inserting the forks at viewing angles \(\psi \gtrsim 10^\circ \) is mechanically infeasible and therefore not a relevant detection scenario. Detections at larger distances and angles are primarily relevant for recognizing the presence of a pallet, whereas closer to the sensor, accurate pose estimation becomes critical. As distance and angle decrease, pose estimation error correspondingly diminishes, as shown in Figure~\ref{fig:pallet_accuracy}. The associated decrease in estimation uncertainty is leveraged in the pallet mapping approach described in Section~\ref{sec:state_estimation}.
Evaluations in Section \ref{sec:experiments} demonstrate that this level of pose accuracy is sufficient to achieve near-human overall system performance. Further details on detection performance are provided in \cite{beleznai_2024}.

\subsection{Loading Edge Detection}
\label{sec:loading_edge_detection}

To accurately position pallets on the truck, the system performs a LiDAR point cloud-based detection of the loading platform, once per load cycle. Given the transformation of the LiDAR sensor relative to a globally planar coordinate frame and a set of 3D points aggregated from LiDAR point clouds over a period of time, the loading platform detector follows a multi-step procedure:

\textbf{Outlier removal:} Outliers are removed by discarding points with insufficient number of neighbors within a specified radius.

\textbf{Height filtering:} The point cloud is refined to include only points within 0.2 and 2.5 meters above the ground.

\textbf{Edge candidate points identification:} Points are classified as edge candidate points if their neighborhood contains two distinct sets of points with orthogonal normals.  One normal direction must align closely with the vertical (up) vector, meaning that the other direction must be parallel to the ground plane.

\textbf{Edge line detection:} A RANSAC (Random Sample Consensus) approach is used to find a line that has strong support in terms of the number of candidate edge points.

\textbf{Loading edge end detection}: The vertical front-face of the loading platform is detected. The loading edge and the front surface together define a reference frame.

\begin{figure}[h]
    \centering
        \includegraphics[width=0.75\linewidth]{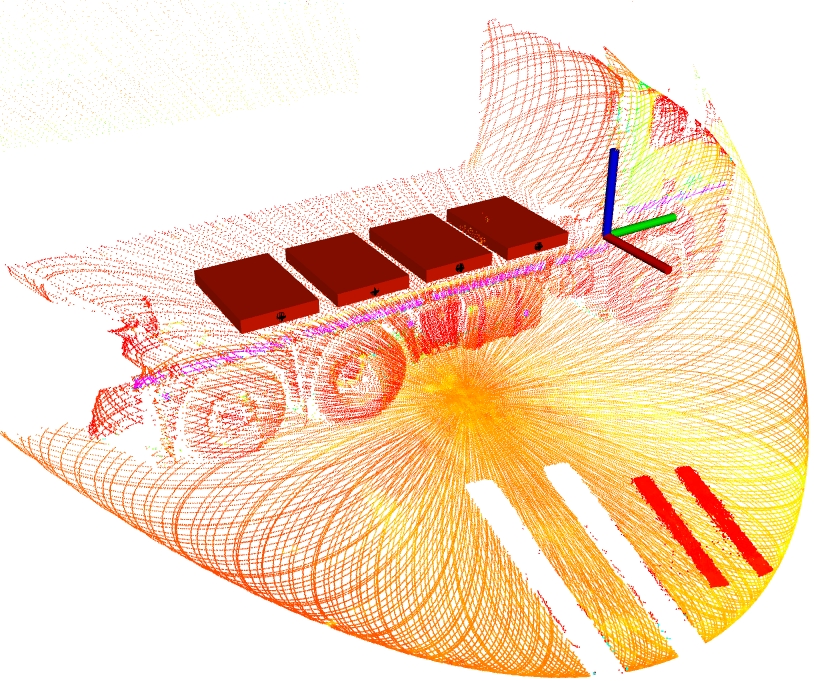}
    \caption{A LiDAR point cloud (colored by reflectance) captures the scene, including the forklift forks (bottom right) and their shadow. The lift is positioned at an estimated loading height to optimize the platform's visibility. The resulting detected loading edge reference frame (RGB frame) and the target slots (red boxes) defined within this frame are depicted as overlay.}
    \label{fig:loading_edge}
\end{figure}

\textbf{Pallet slot definition:} A configurable loading pattern defines target slot placements, depicted in Figure~\ref{fig:loading_edge}.

This process ensures reliable detection of the loading platform, even under challenging conditions, which works regardless of the tilt of the truck or partial loading because it focuses solely on the edge. Additionally, an image segmentation method for part detection was tested \cite{Schwingshackl2024FewShotSegmentation}, but turned out to be more prone to clutter on the platform than this geometry-specific deterministic edge detection.


\subsection{Obstacle Detection}
\label{sec:obstacle_detection}

\begin{figure*}[h!]
    \begin{subfigure}[h!]{0.30\linewidth}
        \centering
        \includegraphics[width=0.6\textwidth]{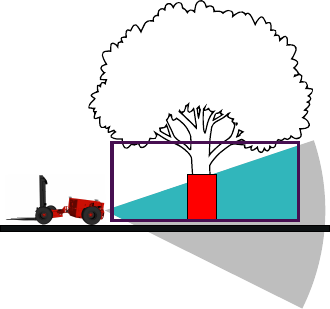}
        \caption{Side-view: collision objects (red) within the sensor field of view (grey) are only detected within a certain height range and distance (cyan). Therefore, the tree's crown is not considered an obstacle.} 
        \label{fig:obstacle_detection_side_view}
    \end{subfigure}
    \hspace{0.04\linewidth}
    \begin{subfigure}[h!]{0.30\linewidth}
        \centering
        \includegraphics[width=0.7\textwidth]{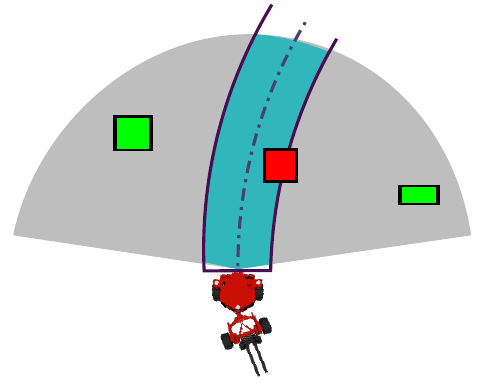}
        \caption{Birds-eye view: Potential obstacles (green) are tracked to assess movement but are only considered as future collisions (red) if they intersect with the path volume (cyan).}
        \label{fig:obstacle_detection_top_down}
    \end{subfigure}
    \hspace{0.04\linewidth}
    \begin{subfigure}[!h]{0.30\linewidth}
        \centering
        \includegraphics[width=\textwidth]{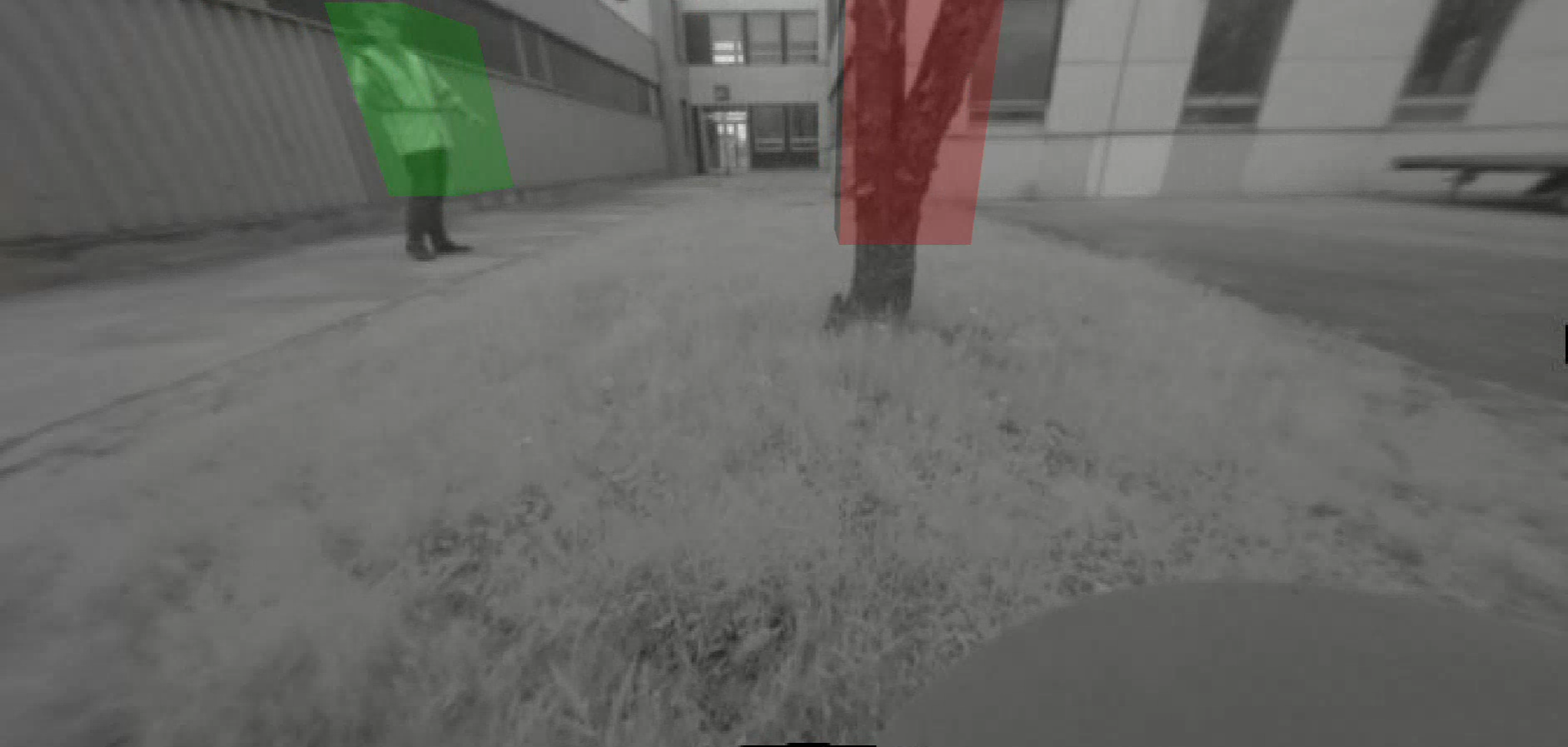}
        \caption{Left stereo camera image with obstacles overlayed. Not all obstacles detected by LiDAR are within the camera's field of view.}
        \label{fig:obstacle_detection_odas_view}
    \end{subfigure}
    \caption{Depiction of the obstacle detection system's geometric relations and output. Definitive collision objects (red) that trigger emergency braking overlap with the expected path volume (cyan), and tracked potential obstacles are outside (green).}
    \label{fig:obstacle_detection}
\end{figure*}

For safe operation, the forklift must detect obstacles quickly to avoid collisions with static obstacles (trees, walls, etc.) and moving obstacles (excavators, pedestrians, etc.), which may approach the currently planned drive path.
The implemented approach is a fast-reacting method in the sensor frame complementary to and independent of the mapping approach presented in Section~\ref{sec:mapping}. It is applied only in the forward driving direction, with a higher risk of colliding with moving obstacles. In contrast, movements in the fork direction, as discussed in Section~\ref{sec:exteroception}, are slower and follow much shorter paths.
Based on the selected safety concept, when an obstacle is detected, the forklift is brought to a safe state (standstill) to prevent collisions, rather than re-planning its path to navigate around the obstacle. Once an emergency stop is triggered, the forklift remains stationary until it receives operator input to resume the task.

For the assessment of the risk of collision with moving obstacles, we utilize a multi-object detection and tracking system, originally based on the works \cite{weichselbaum2013accurate} and \cite{fel2018odas}, which further became the basis for an endurance-tested tramway braking assistance system in commercial use\footnote{https://www.alstom.com/press-releases-news/2026/2/smarter-trams-safer-systems-alstoms-driver-assistance-solutions (accessed 2026-06-03)}. 
It operates in a tracking-by-detection paradigm. 
It can detect arbitrary objects, track them over time, 
and assess if a collision will occur based on the predicted trajectories of the objects and the projected path of the forklift. 
The detection step is based on generic 3D clustering. Hence, it does not depend on predefined object classes such as persons or vehicles.

In order to detect obstacles, the system analyzes point clouds from the Ouster LiDAR mounted on the front part of the forklift. Both LiDAR and a custom stereo camera setup (see Figure~\ref{fig:system_overview}c) have been successfully used, but the wider field of view of the LiDAR system makes it the safer choice for obstacle detection (see Figure~\ref{fig:obstacle_detection}). 
Points above an assumed ground plane (cf. Figure~\ref{fig:obstacle_detection_side_view}) are analyzed for potential obstacles by grouping them into separate distinct objects using Density-Based Spatial Clustering of Applications with Noise (DBSCAN) \cite{ester1996dbscan}.
The clusters represent potential obstacles (indicated as red and green boxes in Figure~\ref{fig:obstacle_detection}). However, only if they overlap with the expected vehicle's path are they considered potential collision objects (red) that trigger an emergency brake.
The volume of the monitored path (cyan in Figure~\ref{fig:obstacle_detection}) is defined by extruding the vehicle footprint (plus an additional clearance buffer) along a simplified vehicle path. 

Obstacle bookkeeping is implemented similarly to pallet bookkeeping, as described in Section~\ref{sec:state_estimation}. However, it is enhanced with a Kalman filter based on a constant velocity model to accommodate movement.
Then each object track is checked for a possible collision, which can cause a collision warning based on the predicted movement of the object, the velocity of the forklift, the characteristics of the forklift deceleration, and a defined minimum safety distance. 

\section{Planning and Control}
\label{sec:planning_control}

\ADAPT performs two primary types of actions to carry out its tasks effectively: (1) collision-free navigation, where the forklift moves from its current position to a target location within a predefined operational area, either while carrying a pallet or traveling unloaded, and (2) precise manipulation, which involves the accurate pick-up and placement of pallets in specified locations (slots), such as on a truck or within a designated area on the ground (multiple slots on a truck are shown in Figure~\ref{fig:loading_edge}).

This section explains how these actions are coordinated throughout the operational workflow and the measures implemented to ensure their accurate and reliable execution.

\subsection{Task Planning and Execution}
\label{sec:task_planning}

\begin{figure}[]
    \centering
    \includegraphics[width=\linewidth]{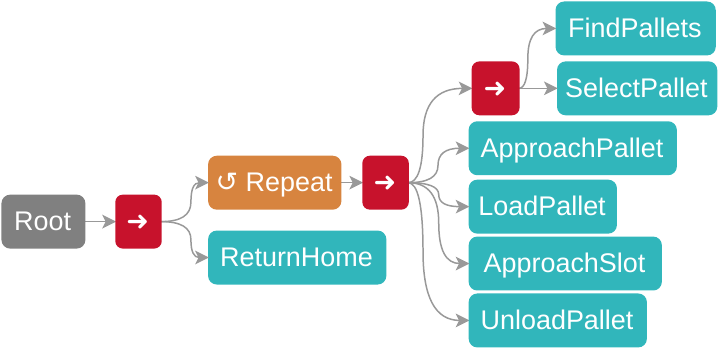}
    \caption{Simplified behavior tree for the proposed system, consisting of the root node (gray), action nodes (cyan), sequences (red) and a decorator node (orange).}
    \label{fig:behavior_tree}
\end{figure}

Task (re-)planning and execution are crucial for the efficient operation of autonomous machines, particularly in dynamic, unstructured environments, enabling them to adapt to changing conditions.
In the proposed autonomous system, task planning and execution are achieved using a behavior tree-based approach. Behavior trees provide a flexible, modular framework for modeling decision-making processes and task execution in a hierarchical, reactive manner \cite{btsurvey2018}. 
A behavior tree consists of action nodes (which perform tasks), condition nodes (which evaluate specific states), and composite nodes (which control the flow of execution, such as sequences and fallbacks). 
This structure allows for scalable and adaptable behavior design, particularly suited for managing complex tasks like navigation and object manipulation.
The integrated behavior tree logic is based on the BehaviorTree.CPP library \cite{btcpp2018}, which offers a performant set of tools for behavior tree construction and execution. 
In order to enable smooth communication with the robot's ROS 2 based architecture, the ROS 2 wrappers from the Nav2 framework \cite{macenski2023survey} are utilized.
A simplified version of the behavior tree that describes the operation of \ADAPT is shown in Figure~\ref{fig:behavior_tree}.
The flow of execution can be summarized as follows:
\begin{compactenum}
    \item The sequence begins with \ADAPT searching for pallets using the \textit{FindPallets} action.
    \item Once pallets are identified, the \textit{SelectPallet} action determines which pallet to target based on predefined criteria.
    \item The \textit{ApproachPallet} action guides \ADAPT in navigating toward the selected pallet.
    \item Once positioned correctly, the \textit{LoadPallet} action ensures the pallet is securely lifted onto \ADAPT.
    \item \ADAPT executes the \textit{ApproachSlot} action to navigate to the designated slot or unloading area.
    \item At the slot, the \textit{UnloadPallet} action places the pallet in its target location.
    \item After completing the unloading task, the behavior tree loops back using the \textit{Repeat} element and is ready for the next pallet. After loading all pallets, the \textit{ReturnHome} action navigates \ADAPT to its home position.   
\end{compactenum}
A more detailed version of the full behavior tree is provided in Appendix~\ref{app:behavior_tree}.
It is considerably more complex, featuring 26 nested sub-trees and more than 30 unique action and composite nodes. 
It integrates both reactive and anticipatory behaviors. 
For example, a reactive behavior occurs after a failed pallet pick-up attempt, due to, for example, inaccuracies in the estimated pose, triggering a recovery action and another pick-up attempt. 

To initiate the operation, \ie~the execution of the behavior tree, \ADAPT receives a command by the forklift supervisor to transport either a specified number of pallets or all available pallets from a designated loading zone to an unloading zone.
The current system iteration features a tablet-based HMI for monitoring and controlling the vehicle's operation. The framework also supports the testing of novel interaction modalities, like gesture control for freeing the hands of an operator; see \cite{Zafari2024} for details. 

\textbf{Operator override and recovery protocol:}
To ensure robust and flexible operation, the system supports seamless transition between autonomous and manual control. The intervention process in case of failure is as follows:
\begin{compactenum}
    \item Errors are indicated through the external HMI (LED). The behavior tree pauses at the current state, and the vehicle remains in a safe emergency stop until the supervisor confirms to continue.
    \item The supervisor assesses the severity of the error and chooses one of two options: proceed with autonomous execution (3a) or switch to manual intervention (3b).
    \item[3a.] To resume autonomous execution, the supervisor presses a button on the certified safety remote. The behavior tree continues from the previously paused action.
    \item[3b.] The supervisor switches to manual mode (via the safety remote) to resolve the error, directly operating all low-level vehicle functions remotely. Once the issue is resolved, the supervisor returns the system to autonomous mode and restarts the task.
\end{compactenum}

\subsection{System Models}
Accurate modeling of the vehicle's kinematics and hydraulic actuators allows precise motion control. This is essential for efficient operation in complex environments, allowing for reliable navigation and pallet manipulation.

\subsubsection{Vehicle Base Kinematics}
\label{sec:articulated_model}

\begin{figure}
    \centering
    \includegraphics[width=0.8\linewidth]{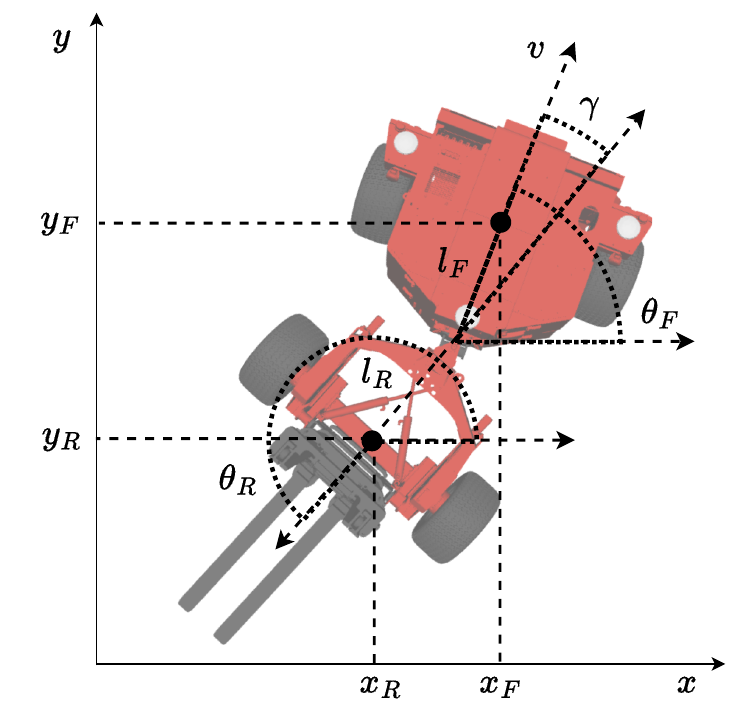}
    \caption{Articulated forklift kinematics with the front part ${\mathbf{q}_{F}} = [x_F, y_F, \theta_F]^\mathrm{T}$, where $x_F$ and $y_F$ represent the position in the 2D plane, and $\theta_F$ its orientation. The rear part is characterized by its own position and orientation  ${\mathbf{q}_{R}} =[x_R, y_R, \theta_R]^\mathrm{T}$. The angle of the articulated steering joint is denoted by $\gamma$.}
    \label{fig:kinematics}
\end{figure}

The two parts of the vehicle chassis are connected by an articulated joint as described in Section~\ref{sec:actuation}. 
This allows modeling the movement on the 2D plane using articulated vehicle kinematics, as shown in Figure~\ref{fig:kinematics}, as
\begin{equation}
\begin{bmatrix}
\dot{x}_F \\
\dot{y}_F \\
\dot{\theta}_F \\
\dot{\gamma}
\end{bmatrix}
=
\begin{bmatrix}
\cos{(\theta_F)} & 0 \\
\sin{(\theta_F)} & 0 \\
\frac{\sin{(\gamma})}{l_F \cos{(\gamma)} + l_R} & \frac{l_R}{l_F \cos{(\gamma)} + l_R} \\
0 & 1
\end{bmatrix}
\begin{bmatrix}
v \\
\dot{\gamma}
\end{bmatrix} \ ,
\label{eq:sys_kin_equations}
\end{equation}
where the Cartesian coordinates of the front axle are defined as $[x_F, y_F]$, with its orientation represented by the angle $\theta_F$.
The rear part ${\mathbf{q}_{R}} =[x_R, y_R, \theta_R]^\mathrm{T}$ is connected to the front part at the articulation point, with its orientation defined by $\theta_R = \pi +\theta_F - \gamma$.
The kinematics are characterized by the system inputs $v$ and $\dot{\gamma}$, which are the velocity at the front axle and the steering rate, respectively.
The parameters $l_F$ and $l_R$ are the center-to-axle distances from the front and the rear axle to the articulation point.

These kinematic equations, which model the motion of the vehicle in the 2D plane, are essential for path planning and vehicle base motion control. Extending this model to account for motion in 3D for steep terrain is an ongoing area of research and a focus of future work.

\subsubsection{Hydraulic Modeling}
\label{sec:hydraulic_model}

\begin{figure*}[h!]
    \centering
    \begin{subfigure}{0.196\textwidth}        
        \centering
        \caption{Drive}         
        \includegraphics[width=\textwidth]{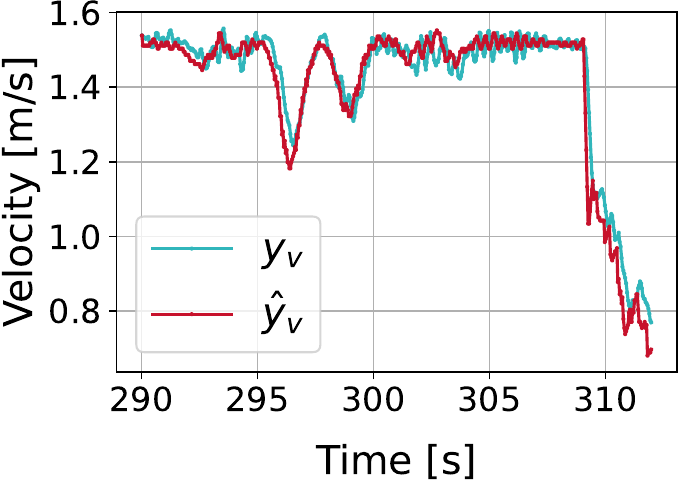}               
        \label{fig:model_drive}        
    \end{subfigure} 
    \begin{subfigure}{0.196\textwidth}        
        \centering
        \caption{Steer}
        \includegraphics[width=\textwidth]{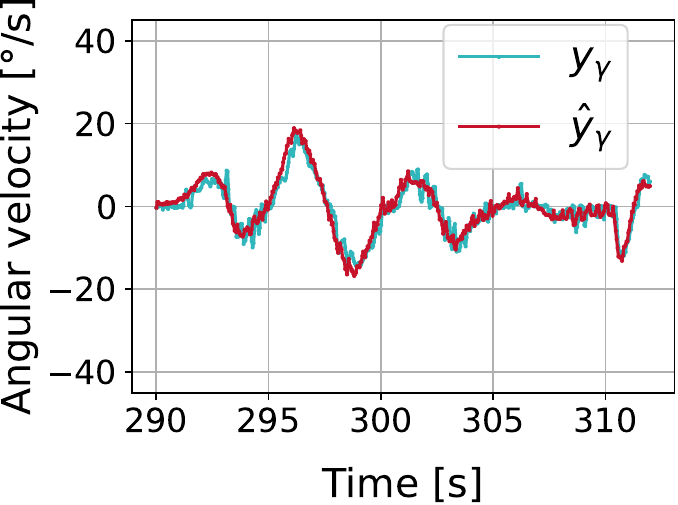}             
        \label{fig:model_steer}
    \end{subfigure} 
    \begin{subfigure}{0.196\textwidth}       
        \centering
        \caption{Tilt}
        \includegraphics[width=\textwidth]{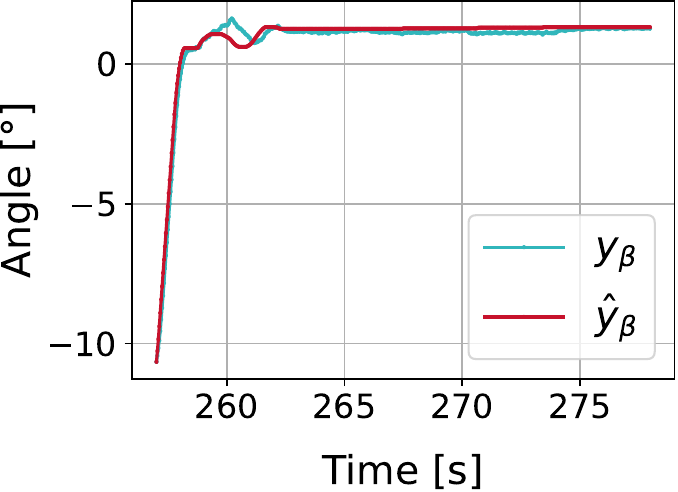}        
        \label{fig:model_tilt}
    \end{subfigure}  
    \begin{subfigure}{0.196\textwidth}        
        \centering
        \caption{Lift}
        \includegraphics[width=\textwidth]{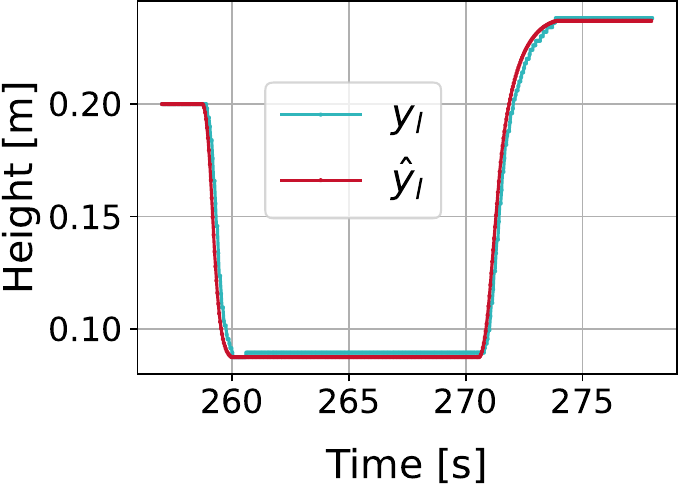}        
        \label{fig:model_lift}
    \end{subfigure} 
    \begin{subfigure}{0.196\textwidth}        
        \centering
        \caption{Shift}
        \includegraphics[width=\textwidth]{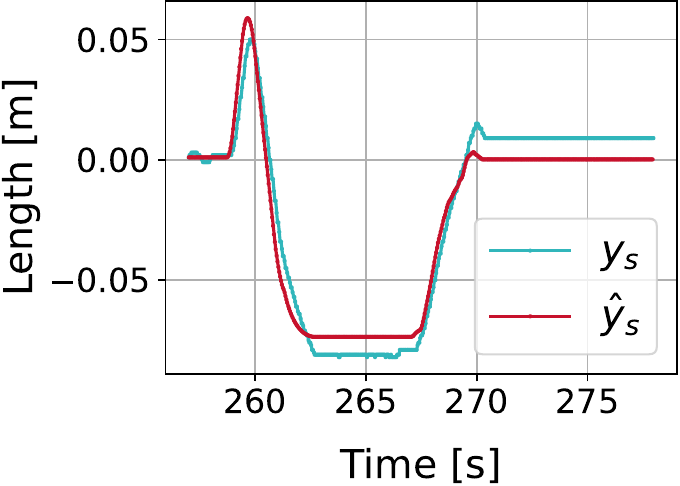}        
        \label{fig:model_shift}
    \end{subfigure}  
    \vspace{-0.3cm}
    \caption{Results for modeling the hydraulic drive and the hydraulic cylinders showing the measured system output $y_i$ compared to the estimated system output $\hat{y_i}$, calculated from the actuation command with $i \in \{v,\gamma,\beta,l,s\}$. Subfigures (a) and (b) show data from a path following scenario, and Subfigures (c)-(e) refer to a pallet pick-up maneuver.}
    \label{fig:cylinder_models}
\end{figure*}

The flow rate of the hydraulic fluid in the hydrostatic four-wheel drive system is determined by the duty cycle of a PWM signal $u_{v}$, which has a nearly linear relation to the wheel velocity $y_v$. This is modeled by 
\begin{equation}
    \hat{y}_v(t) = K_1 u_{v}(t) + K_0 \ , 
    \label{eq:linear_drive}
\end{equation}
with the estimated velocity $\hat{y}_v$. The parameterization of the constants $K_0$ and $K_1$ was empirically determined by identification runs.
Figure~\ref{fig:model_drive} shows the estimated velocity $\hat{y}_v$ compared to the measurement $y_v$. 

The remaining degrees of freedom, namely steering ($\gamma$), tilting ($\beta$), lifting ($l$), and shifting ($s$), are actuated by hydraulic cylinders. 
The angular velocity of the steering joint $\hat{y}_\gamma$ can be modeled as a linear function of the desired valve spool position of the steering cylinder $u_\gamma$, similar to \eqref{eq:linear_drive}.
The other cylinders are modeled as first-order lag elements 
\begin{equation}
    \tau_\zeta \dot{\hat{y}}_\zeta(t) + \hat{y}_\zeta(t) = K_\zeta u_\zeta(t) \ ,
    \label{eq:pt1}
\end{equation}
where the input $u_\zeta$ represents the desired valve spool position for the corresponding actuated cylinder and $\hat{y}_\zeta$ denotes the estimated output of the system, $K_\zeta$ a model constant, and $\tau_\zeta$ a time constant for the model of $\zeta \in \{\beta,l,s\}$ tilting, lifting, and shifting, respectively. 
The results of the parameter identification experiments are shown in Figure~\ref{fig:cylinder_models}, applied to a path following (Figure~\ref{fig:model_steer}) and a pallet pick-up scenario (Figure~\ref{fig:model_tilt}-\ref{fig:model_shift}).
The estimated cylinder motion, denoted as $\hat{y}_\zeta$, aligns closely with the measured variable $y_\zeta$. However, when multiple cylinders are actuated simultaneously, the model's accuracy decreases due to altered pressure distribution, flow division, and pump limitations, leading to nonlinear behavior. This reduction in accuracy is not significant in operation, as it can be compensated for by the feedback control system, discussed in the next section.

\subsection{Motion Planning and Control}
\label{sec:motion_control}

The operation of \ADAPT is divided into two primary action types: (1) Navigation between poses, comprising path planning and following, and (2) manipulation for loading or unloading pallets. 
These primary actions share common vehicle functions, \ie vehicle base and fork positioning control.
These functions are implemented as two separate cascaded control structures, as illustrated in Figure~\ref{fig:control_architecture}. The main components of these control loops are the pose tracking control (PTC) and fork tip transformation
(FTT), which track the external reference. The inner control loop consists of a feedforward control (FF) based on the hydraulic models from Section~\ref{sec:hydraulic_model} and basic PI(D) control and saturation blocks for hydraulic valve and cylinder control.
This section delves into the details of the primary action types and explains how vehicle functions are implemented. 

\subsubsection{Cascaded control}
\label{sec:controls_structures}

\begin{figure}
\centering     
    \includegraphics[width=1.0\linewidth]{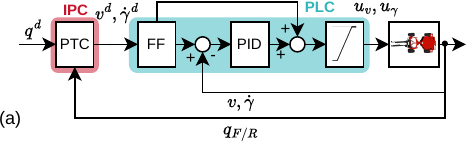} 
    \includegraphics[width=1.0\linewidth]{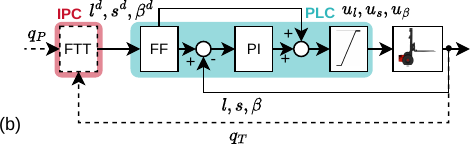}
    \caption{Cascaded control architectures for the vehicle base (a) and fork actuation (b) form the foundation for pose-to-pose navigation and pallet manipulation. The main components, enclosed in red boxes, are the pose tracking control (PTC) and fork tip transformation (FTT), which run on the IPC, while those in cyan boxes operate on the real-time PLC. For navigation, FTT is not needed.}
    \label{fig:control_architecture}
\end{figure}

\textbf{Vehicle base control:}
The algorithm to control the movement of the vehicle base is based on the kinematics of the articulated vehicle described in Section~\ref{sec:articulated_model}.
It is inspired by the work presented in \cite{bian2017kinematics}, which introduces a path-following strategy for articulated drum rollers that navigate straight paths through construction sites. 
The core concept involves tracking a virtual reference vehicle with coordinates ${\mathbf{q}^d} =[x^d, y^d, \theta^d]^\mathrm{T}$, which serve as input of the Pose Tracking Control (PTC) component in Figure~\ref{fig:control_architecture}a. The reference pose ${\mathbf{q}^d}$ can either represent a pose on a precomputed path or a pallet being approached. The exact methods for determining ${\mathbf{q}^d}$ will be explained in the following sections.

The tracking is implemented by minimizing the state of the error system 
\begin{equation}
\begin{bmatrix}
e_x \\ 
e_y \\ 
e_\theta
\end{bmatrix}
=
\begin{bmatrix}
\cos{(\theta^d)} & \sin{(\theta^d)} & 0 \\ 
-\sin{(\theta^d)} & \cos{(\theta^d)} & 0 \\ 
0 & 0 & 1
\end{bmatrix}
\begin{bmatrix}
x_{F/R} - x^d \\ 
y_{F/R} - y^d \\ 
\theta_{F/R} - \theta^d
\end{bmatrix} \ ,
\label{eq:error_system}
\end{equation}
which is calculated by transforming the Cartesian coordinates of the reference vehicle into the vehicle's body-fixed origin ${\mathbf{q}_{F/R}} =[x_{F/R}, y_{F/R}, \theta_{F/R}]^\mathrm{T}$. This can either be the center of the front (F) or the rear (R) axle.
The state comprises the relative errors in longitudinal direction $e_x$, lateral direction $e_y$, and the heading $e_\theta$.
The approach neglects $e_x$ and instead focuses on minimizing $e_y$ and $e_\theta$ which results in the control law for the steering rate 
\begin{equation} \dot{\gamma}^d = -\frac{K_1 v (l_F + l_R)}{l_R} e_y - \frac{K_2 (l_F + l_R)}{l_R} e_\theta - \frac{v}{l_R} \gamma \ ,
\label{eq:gamma_dot_desired}
\end{equation} derived using Lyapunov theory, as detailed in \cite{bian2017kinematics}. 
$K_1$ and $K_2$ are control gains and determine the influence of the respective lateral and heading errors on the desired steering rate.
The second input to the kinematic system is the desired forward velocity 
\begin{equation}
v^d = \max(v_R - k_{v}(\dot{\gamma}^d)^{2},v_{\min}) \ ,
\label{eq:v_desired}
\end{equation}
which is designed to respect the constrained system's steering dynamics. Here, $v_R$ is the desired reference velocity for operation.
The parameter $k_{v}$ allows the reduction of the velocity proportional to the squared desired steering rate. Thus, sharp curvatures are tracked at a lower speed to maintain tracking accuracy.
The minimum velocity $v_{\min}$ ensures that the vehicle maintains a minimum speed, even during sharp steering maneuvers. 
Equations \eqref{eq:error_system}-\eqref{eq:v_desired} build the foundation of the PTC and, thus, the cascaded control loop in Figure~\ref{fig:control_architecture}a. 

\textbf{Fork positioning control:}
For a successful pallet manipulation, an accurate positioning of the fork tip is necessary. The most critical is pallet pick-up, where the tolerated positioning error in lateral and vertical direction must not exceed 5 cm. To this end, the tip of the fork $\mathbf{q}_T = [x_T,y_T,\theta_T,z_T]^\mathrm{T}$ must be precisely aligned with the pallet front $\mathbf{q}_P = [x_P,y_P,\theta_P,z_P]^\mathrm{T}$ (see Figure~\ref{fig:pallet_manipulation_gesamt}). 
This is ensured by the fork tip transformation (FTT) component, as part of the cascaded control loop depicted in Figure~\ref{fig:control_architecture}b. 
The FTT transforms the target pose into the fork tip coordinate frame by

\begin{equation}
\begin{bmatrix}
x_P' \\ 
y_P'
\end{bmatrix}
=
\begin{bmatrix}
\cos{(e_\theta)} & -\sin{(e_\theta)}\\ 
\sin{(e_\theta)} & \cos{(e_\theta)}
\end{bmatrix}
\begin{bmatrix}
x_P - x_T \\ 
y_P - y_T
\end{bmatrix} \ , 
\label{eq:cont_pallet_pose}
\end{equation}
where $e_\theta = \theta_P-\theta_T$.

Based on \eqref{eq:cont_pallet_pose}, the reference signal for the fork shift corresponds to the lateral error, $s^d = y_P'$, and the lift mast height to the vertical error, $l^d = z_P$. 
The tilting command $\beta^d$ is adjusted to keep the forks parallel to the ground.
The control loop relies on the vehicle base control, which minimizes the orientation error \(e_\theta\). If \(e_\theta\) or \(y'_P\) exceed a predetermined threshold shortly before the forks enter the pallet (i.e., \(\epsilon_x > x'_P > 0\), $\epsilon_x=0.2$m), the pallet pick-up procedure is aborted, and a recovery maneuver is initiated via the task planner.

The cascaded loop in Figure~\ref{fig:control_architecture}b can operate without the FTT, depending on the forklift's current task. 
Specifically, during transporting a pallet - while not loading or unloading - the FTT becomes optional, allowing the setpoints for the underlying control loop $l^d,s^d$, and $\beta^d$ to be directly assigned to a transport position that keeps the fork at a safe height and angle.

\subsubsection{Navigation}
The navigation action comprises planning and following a collision-free path from one pose to another.
Planning a path is a well-established research area for robotic applications, with numerous feasible algorithms. Based on an empirical evaluation of potential path planners, the Hybrid A* is advantageous for flexible path planning in construction environments. This is due to its ability to plan near-optimal bi-directional paths while maintaining computational efficiency, and it places no restrictions on path length or the number and location of driving direction changes. Moreover, an adjustable penalty term can be employed to prefer specific driving directions.

\begin{figure}
    \begin{tikzpicture}
        \centering
        \draw (0, 0) node[inner sep=0] {\includegraphics[width=\linewidth]{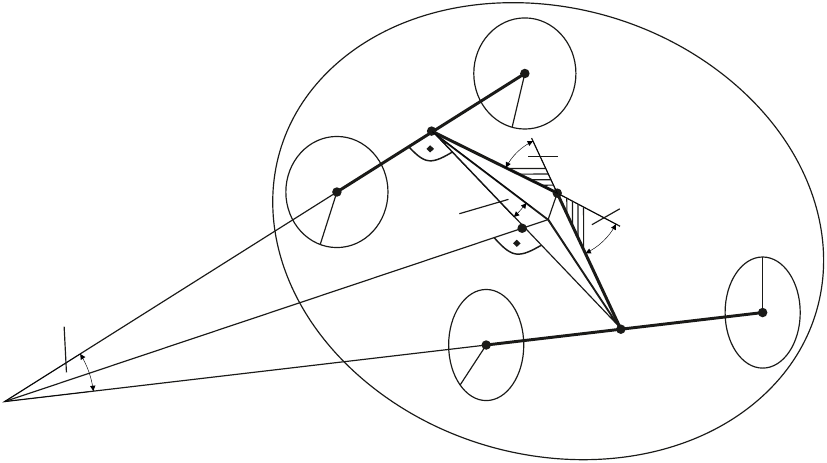}};
        \draw (-4.2,-2.0) node {\footnotesize{ICR}};
        \draw (0.7,1.05) node {\footnotesize{$l_F$}};
        \draw (2.3,-0.5) node {\footnotesize{$l_R$}};
        \draw (0.2,0.1) node {\footnotesize{$\gamma/2$}};
        \draw (-3.9,-1.0) node {\footnotesize{$\gamma$}};
        \draw (-0.7,-0.9) node {\footnotesize{$r_v$}};
        \draw (1.7,0.8) node {\footnotesize{$\alpha$}};
        \draw (2.3,0.3) node {\footnotesize{$\beta$}};
    \end{tikzpicture}
    \caption{ICR for the articulated vehicle in a tilted position: Horizontal shading marks the angle in the horizontal plane and vertical shading in the vertical plane.}
    \label{fig:plan_ICR}
\end{figure}

\begin{figure} 
    \centering
    \includegraphics[width=1.0\linewidth]{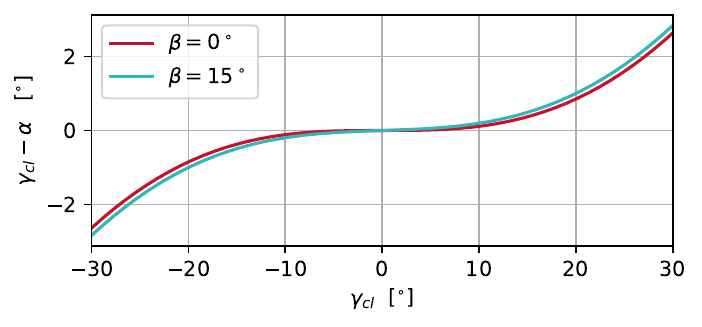}
    \vspace{-15pt}
    \caption{Deviation between car-like steering angle $\gamma_{cl}$ and the articulation angle $\alpha$.}
    \label{fig:plan_steering}
\end{figure}

However, the Hybrid A* algorithm is formulated for car-like vehicles and cannot be directly applied to articulated vehicles. Nonetheless, for symmetrical articulated vehicles, \ie $l_R = l_F$, the kinematic properties for static curvature and straight-line driving are equal to those of car-like vehicles. 
This is achieved by superimposing the Instantaneous Centre of Rotation (ICR), \ie the intersection point of the front and rear axles, of both vehicle models. Figure~\ref{fig:plan_ICR} illustrates the articulated vehicle model with its ICR, the articulation angle $\alpha$, and the tilting angle $\beta$.
Using the geometric relations of the front part, we obtain  the steering angle of the articulated vehicle
\begin{equation}
    \gamma = 
    \arctan{\left(\frac{l_R}{2 r_v} \sqrt{2 \left(\cos\left(\alpha\right) \cos\left(\beta\right) + 1\right) } \right)}
    \label{eq_plan_alpha_arti_1}
\end{equation}
with the radius of the ICR $r_v$. For the car-like model, the same radius of the ICR results in
\begin{equation}
    r_v = \frac{l_R}{2 \sin\left(\gamma_{cl}/2\right)},
    \label{eq:plan_rv_car_like}
\end{equation}
with the steering angle of the car-like model $\gamma_{cl}$. The relation between the steering angle and the articulation angle $\alpha$ follows from the rotation between the front and rear part
\begin{equation}
    \gamma = 
    \arccos{\left(\sqrt{\frac{\sin^2\left(\alpha\right) \left(\sin^2\left(\beta\right) - 1\right)}{2\left(\cos\left(\alpha\right) \cos\left(\beta\right) + 1\right)}} \right)}.
    \label{eq_plan_alpha_arti_2}
\end{equation}
These two equations \eqref{eq_plan_alpha_arti_1} and \eqref{eq_plan_alpha_arti_2}, together with \eqref{eq:plan_rv_car_like}, can be solved numerically for the steering angle $\gamma$ or the articulation angle $\alpha$. Figure~\ref{fig:plan_steering} shows the deviation between the planned angle $\gamma_{cl}$ for car-like steering and the kinematic-matching articulation angle~$\alpha$.

\begin{figure}
    \centering    
    \includegraphics[width=0.8\linewidth]{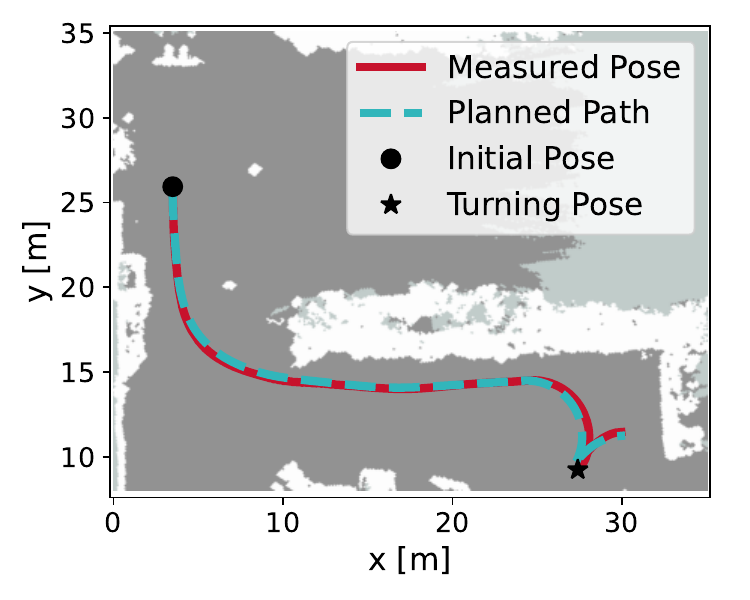}  
    \caption{Bi-directional path planned (cyan), together with path tracking results (red) including one turning pose ($\bigstar$). White areas indicate obstacles.}
    \label{fig:pfc_bidirectional}
\end{figure}

The Hybrid A* implementation used in this work was developed in C++ using ROS 2 and is described in detail in \cite{macenski2024smac}. Figure~\ref{fig:pfc_bidirectional} illustrates an exemplary path planning result for bi-directional movement. The occupancy map utilized for collision-free planning is generated using the algorithms detailed in Section~\ref{sec:mapping}.

\begin{figure*}[h]
    \begin{subfigure}[t]{0.34\linewidth}
        \centering
        \includegraphics[width=1.0\textwidth]{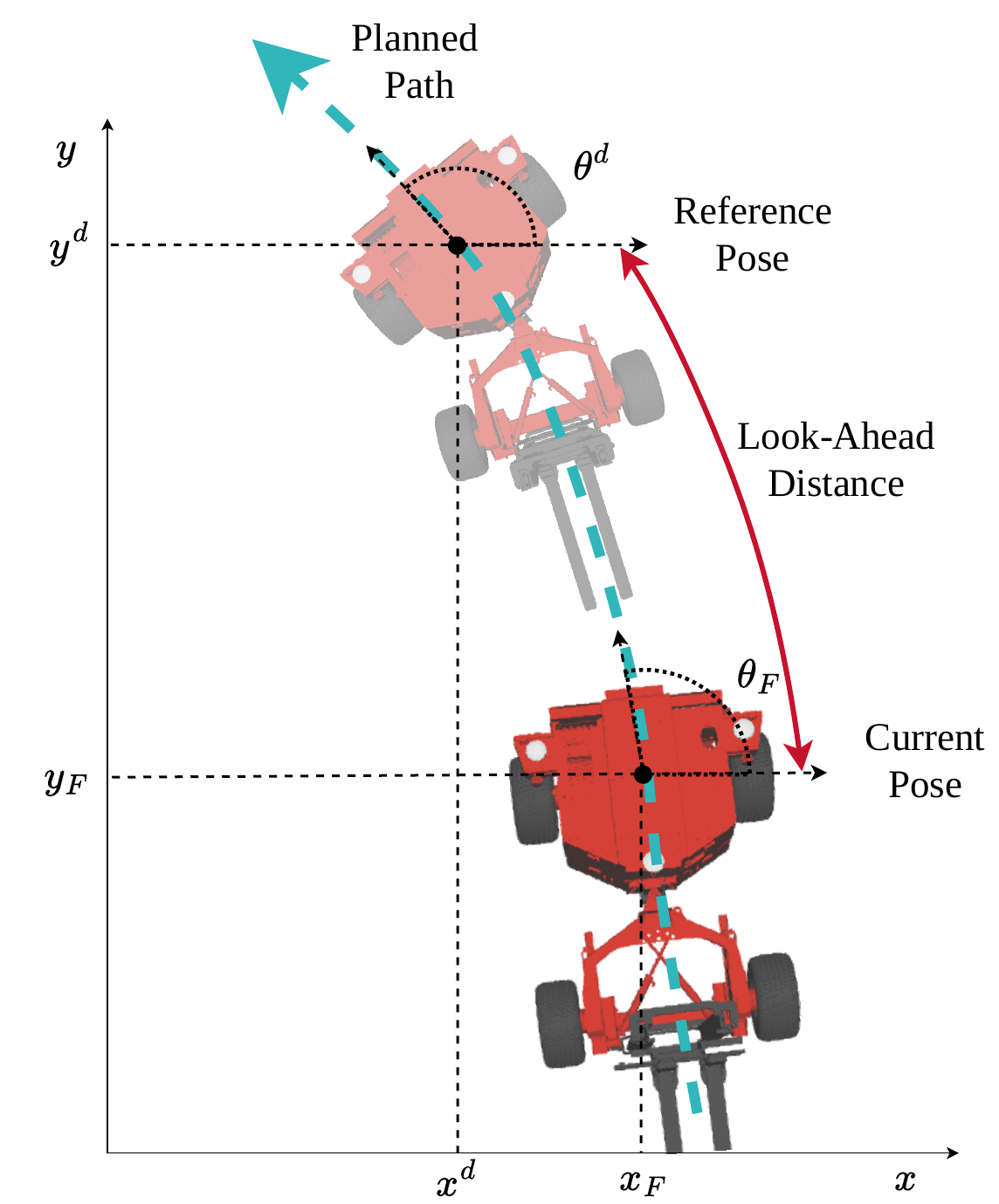} 
        \caption{Concept for path following, showing the virtual vehicle $\mathbf{q}^d = [x^d, y^d, \theta^d]^\mathrm{T}$ as the reference pose.}
        \label{fig:path_following}
    \end{subfigure}
    \hspace{0.02\linewidth}
    \begin{subfigure}[t]{0.63\linewidth}
        \centering
        \includegraphics[width=1.0\textwidth]{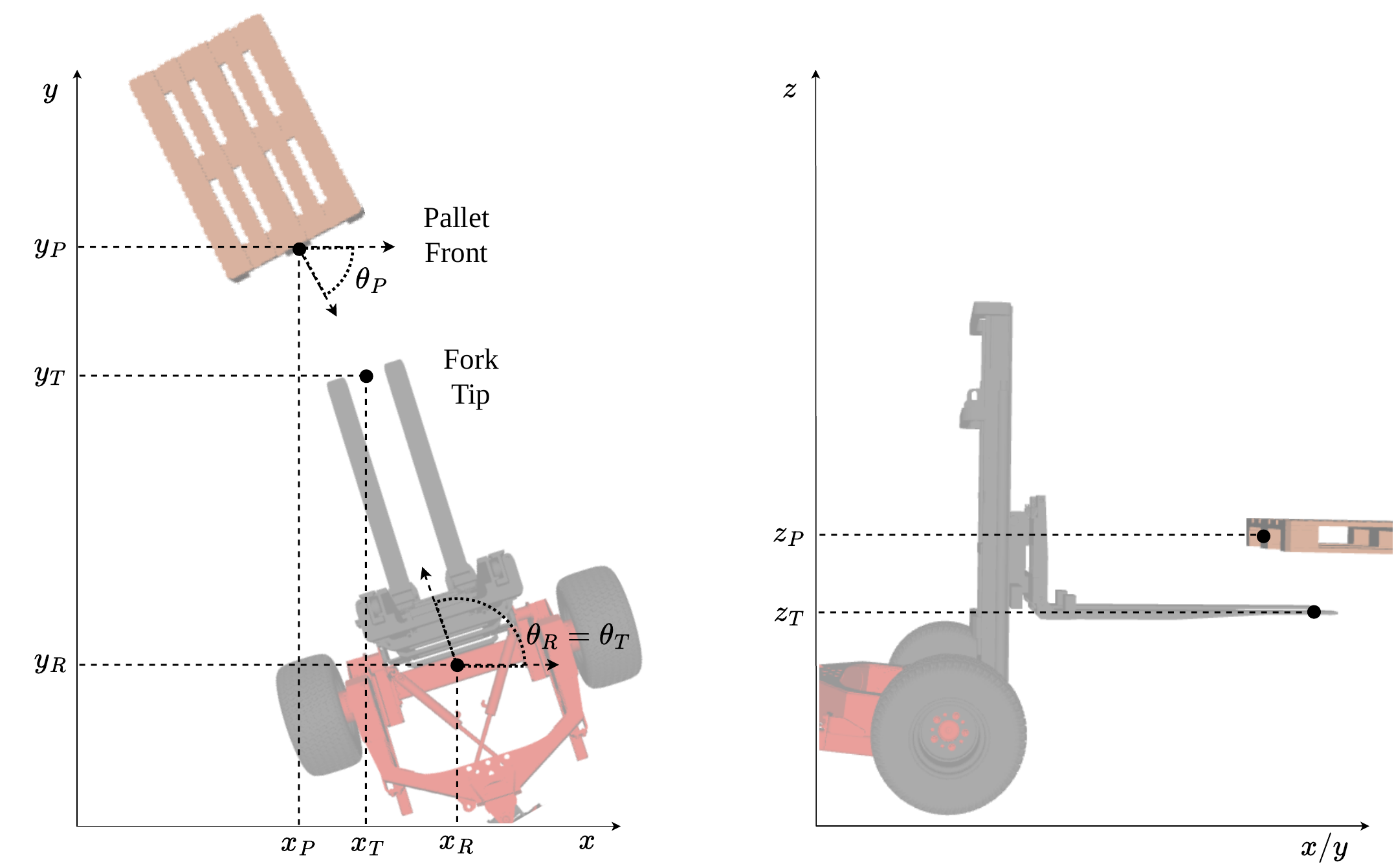}  
        \caption{Pallet manipulation scenario, showing the rear of the vehicle ${\mathbf{q}_{R}} =[x_{R}, y_{R}, \theta_{R}]^\mathrm{T}$, the pallet pose $\mathbf{q}_P =[x_P, y_P, \theta_P, z_P]^\mathrm{T}$ and the tip of the forks $\mathbf{q}_T =[x_T, y_T, \theta_T,z_T]^\mathrm{T}$.}
        \label{fig:pallet_manipulation_gesamt}
    \end{subfigure}
    \caption{Concepts for path following and pallet manipulation.}
\end{figure*}

The subsequent path following ensures precise navigation along the planned collision-free path, effectively avoiding obstacles for safe operation. The core idea, illustrated in Figure~\ref{fig:path_following}, is to track a virtual reference vehicle moving along the planned path with coordinates ${\mathbf{q}^d} =[x^d, y^d, \theta^d]^\mathrm{T}$, which corresponds to the input of the vehicle base control structure described in \eqref{eq:error_system}-\eqref{eq:v_desired}.
The position of the virtual reference vehicle is calculated by identifying the closest pose on the path to the actual vehicle pose ${\mathbf{q}_{F/R}} =[x_{F/R}, y_{F/R}, \theta_{F/R}]^\mathrm{T}$, depending on forward (F) or reverse (R) driving and, from this pose on, simulating a movement along the path for a specified look-ahead distance.
This distance determines how far ahead the vehicle anticipates along the path, balancing responsiveness and stability for smooth and accurate path following.
The pose resulting in the path after this simulated movement serves as the reference pose ${\mathbf{q}^d}$ for tracking the path.

Since the forklift is designed to operate in both forward and reverse driving directions and the implemented path tracking control law is designed for uni-directional movement, the bi-directional path following approach employs a multi-stage approach.
The bi-directional path consists of segments, separated by turning poses, with alternating movement direction. Therefore, the respective vehicle pose also alternates between the front $\mathbf{q}_{F}$ and the rear axle $\mathbf{q}_{R}$.
This approach is facilitated by the forklift's symmetric geometry and the behavior tree handles smooth transitions between the segments.
An example of bi-directional path following, including a turning pose, is shown in Figure~\ref{fig:pfc_bidirectional}.

In addition to controlling the vehicle base, the path-following module utilizes the simplified fork control loop in Figure~\ref{fig:control_architecture}b without FTT to ensure that the forks maintain a safe height and angle, particularly when transporting a load. This is achieved by directly applying the predefined setpoint values $l^d, s^d$, and $\beta^d$ for the respective joints.

\subsubsection{Manipulation}
The manipulation action involves the correct pick-up and drop-off of the pallets. The high-level task planning framework controls this action by selecting a suitable target pallet based on spatial proximity and accessibility and executing a controlled approach.
The vehicle base and fork cascaded control loops from Section~\ref{sec:controls_structures} are necessary to ensure correct pallet pick-up.
The aim is to minimize the error between the pallet pose and the vehicle's rear axle.
The target pose for the vehicle base control is now the center of the front face of the pallet ($\mathbf{q}^d = \mathbf{q}_P$), and the vehicle origin is the center of the rear axle $\mathbf{q}_R$\footnote{Remember that the forks are mounted at the rear of the vehicle.}, as illustrated in Figure~\ref{fig:pallet_manipulation_gesamt}.

\begin{figure}
    \centering
    \includegraphics[width=1\linewidth]{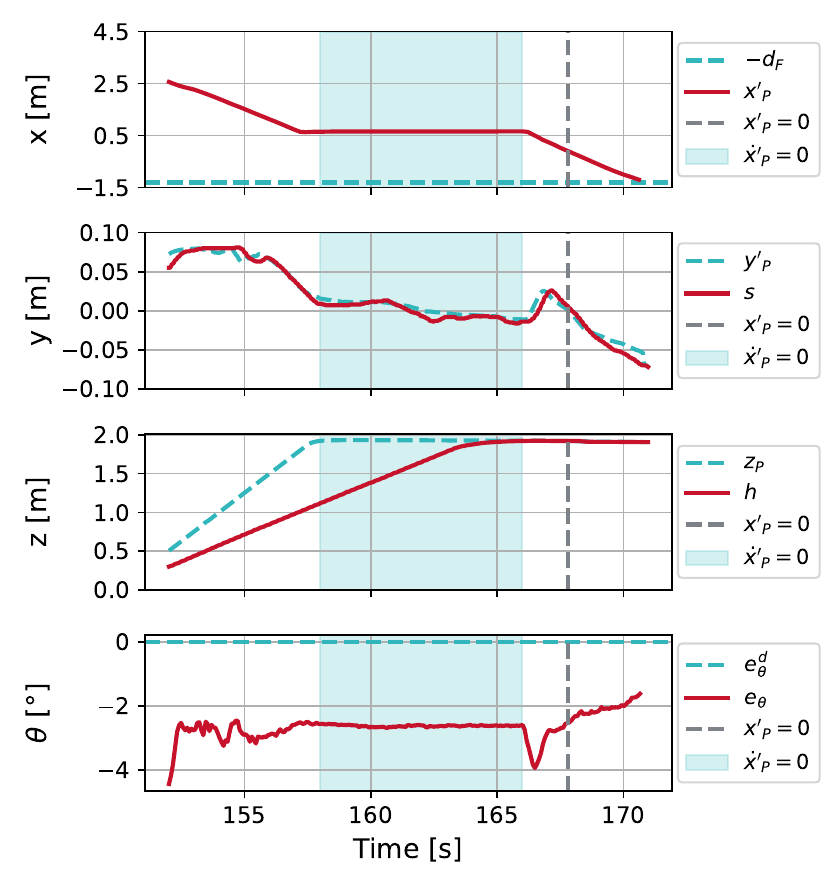}
    \vspace{-22pt}
    \caption{Exemplary data for placing a pallet in a slot on the truck. Cyan lines indicate setpoint signals, red lines refer to measurements, and the gray line marks the time when the fork tip reaches the front of the slot. The standstill time ($\dot{x}'_P=0$) of $\approx10$ seconds results from the waiting time required for the forks to reach the desired height.}
    \label{fig:pallet_putdown}
\end{figure}

During the pallet approach phase, the FTT component from Figure~\ref{fig:control_architecture}b is crucial to align the fork tips with the pallet openings, before and while the vehicle is moving onto the pallet.
The reverse action of placing the pallet follows a process similar to pick-up. 
Here, a virtual pallet called \textit{slot} defines the target location, guiding the system to the desired position.
Figure~\ref{fig:pallet_putdown} illustrates data from a successful pallet manipulation maneuver, showing the process of placing a pallet into a slot on the truck. 
The critical moment for loading the pallet, highlighted by a dashed gray line, occurs when the fork tips reach the front of the (virtual) pallet, and the forks must properly engage with the pallet pockets. 
The vehicle stops moving ($v^d=0$) at $x_P'=-d_F$, where $d_F$ corresponds to the length of the forks or if the fork-mounted laser sensor indicates successful pallet insertion.

Initial iterations of ADAPT have shown that placing the pallet on its designated slot purely by estimating its z-position based on the ground plane or a truck loading edge is not sufficiently robust. Thus, we introduce a simple improvement procedure based on measuring the hydraulic pressure in the fork-lifting cylinder, as follows:
\begin{compactenum}
    \item The forklift enters the selected slot at a safe height above the estimated pose.
    \item After reaching the desired pose in x and y, the vehicle stops.
    \item The forks are slowly lowered until the hydraulic pressure of the lifting cylinder drops, indicating that the fork is in contact with the underlying ground.
    \item Finally, the fork is lifted again by half the height of the pallet to safely retract from the pallet.
\end{compactenum}
The method for estimating the fork contact $f_c$ is explained in Algorithm~\ref{alg:fork_contact}.
\begin{algorithm}[h]
\caption{Fork contact detection.}
\label{alg:fork_contact}    
    \textbf{Input:} Lifting pressure $p_l$. \\
    \textbf{Output:} $f_c$: $true$ if fork in contact, $false$ otherwise.
    \begin{algorithmic}[1]        
       \If {$p \leq p_{pc}$}
           \State $crit_{cnt}\leftarrow crit_{cnt}+1$,  $f_{pc}\leftarrow$$true$
        \Else        
            \State $crit_{cnt}\leftarrow0$, $f_{pc}\leftarrow$$false$
        \EndIf        
        \If {$crit_{cnt} > crit_{tout}$ \textbf{or} $p_l < p_{c}$}
            \State $f_c\leftarrow$$true$, $f_{pc}\leftarrow$$true$
        \Else
            \State $f_c\leftarrow$$false$
        \EndIf
    \end{algorithmic}
\end{algorithm}

\begin{figure}
    \centering
    \includegraphics[width=1.0\linewidth]{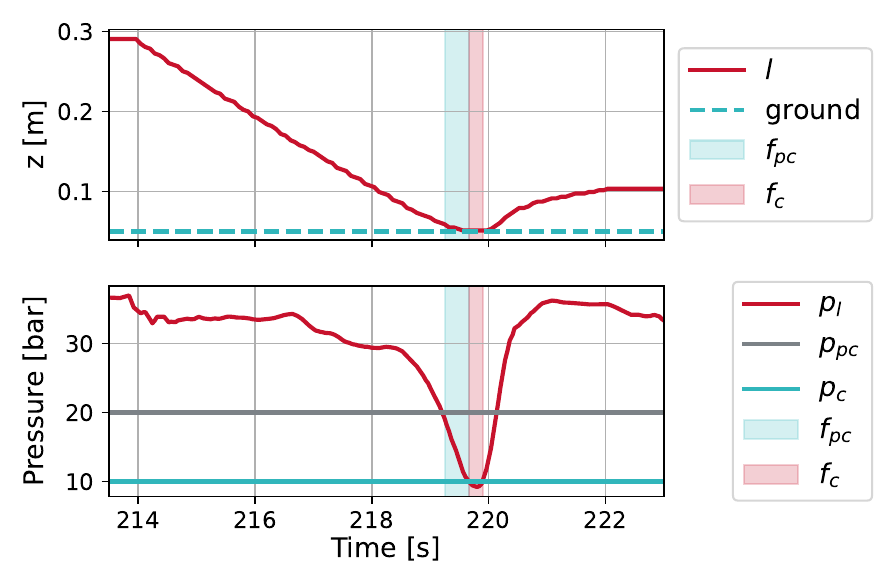}
    \vspace{-22pt}
    \caption{Fork contact detection. Hydraulic pressure in the lifting cylinder $p_{l}$ below a certain level indicates fork contact with environment ($f_{c} = true$). The threshold levels indicate possible fork contact ($p_{pc}$) and definite fork contact ($p_{c}$).}
    \label{fig:fork_contact}
\end{figure}

A graphical description based on exemplary pallet unloading measurements is shown in Figure~\ref{fig:fork_contact}. In addition to ensuring that a pallet is placed at the correct height, this method also serves as a safety feature to prevent dangerous downward fork movement in case the forks come into contact with the environment. The parameters indicating possible ($p_{pc}$) and definite fork contact ($p_{c}$,$crit_{tout}$) are empirically selected as a trade-off between operational robustness and safety.

\section{Validation and Evaluation}
\label{sec:experiments}
This section describes the experimental setup in our outdoor large-scale robotics lab followed by a detailed analysis of the robust continuous operation together with a direct comparison with an experienced human operator. 

\subsection{Experimental Setup}
 
\begin{figure}
    \centering
    \includegraphics[width=0.8\linewidth]{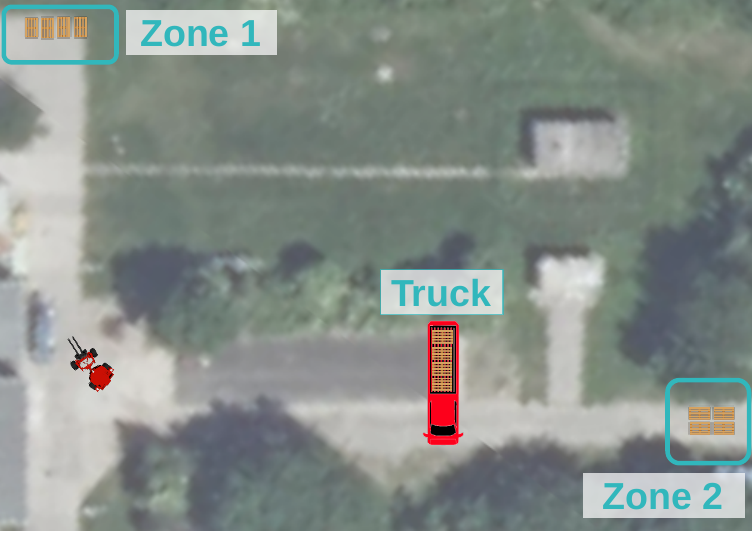}
    \caption{Satellite image showing the test site with an overlay for the configuration of the loading zones (Zone 1 \& Zone 2) and the truck location used for the experiments.}
    \label{fig:testsite_setup}
\end{figure}

The experiment is designed to evaluate the operational robustness and performance of \ADAPT across close-to-real loading scenarios. Three distinct operational modes were tested: ground-to-ground (G to G) pallet loading, ground-to-truck (G to T) pallet loading, and truck-to-ground (T to G) pallet unloading. In this context, the experimental setup involved a truck loaded with four Euro-pallets, each carrying varying types of loads to simulate real-world handling conditions. 
The initial location for the forklift and the truck parking location are unchanged in all experiments, to allow for a fair comparison between baseline data collected from an expert operator and data collected from autonomous operation. Figure~\ref{fig:testsite_setup} shows the test site with an overlay of the configuration of the ground loading zones (Zone 1 \& Zone 2) as well as the truck location. 

\begin{figure}
    \centering
    \includegraphics[width=0.48\linewidth]{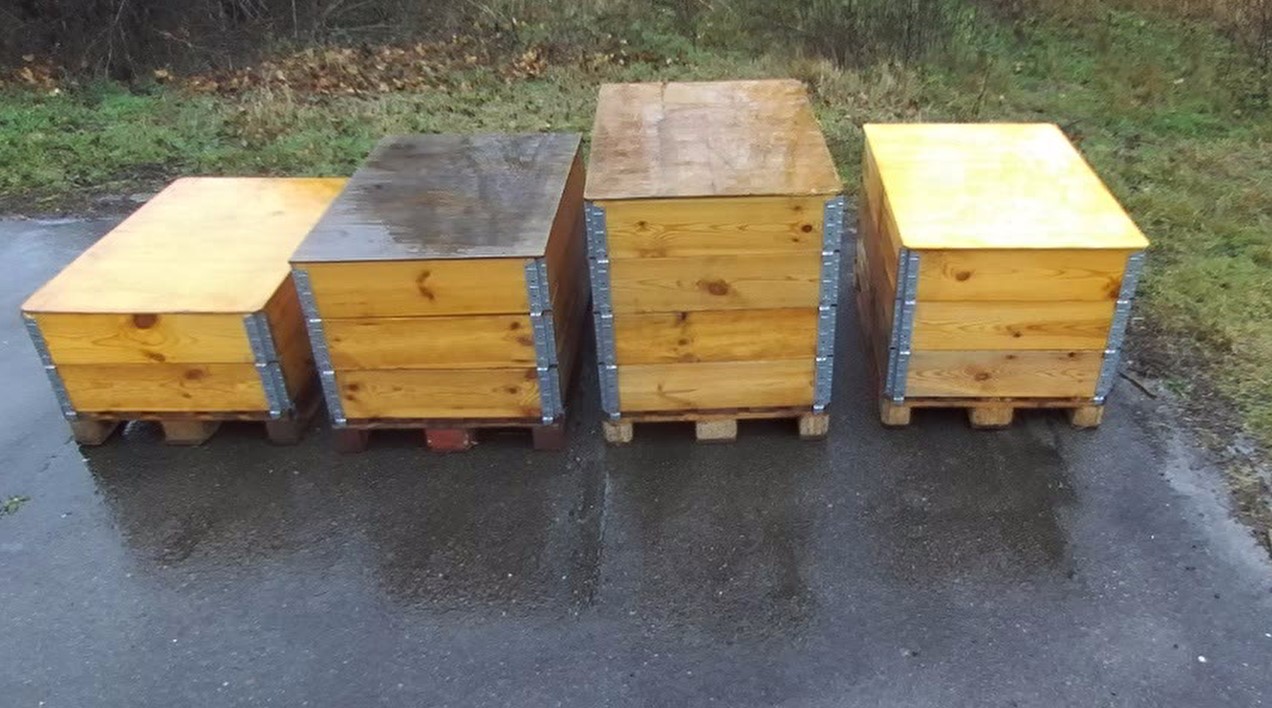}
    \includegraphics[width=0.46\linewidth]{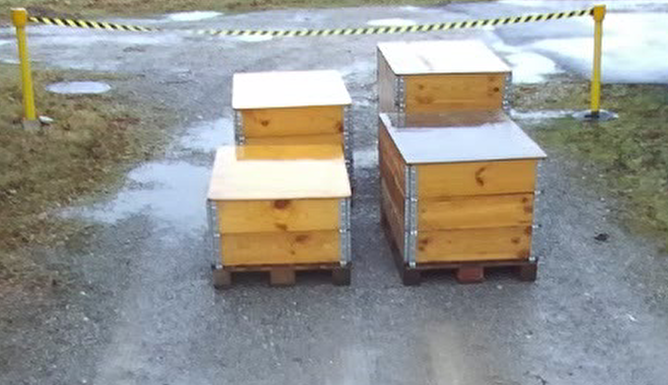}
    \caption{Pallets from the camera on top of the lift mast for Zone 1 (left) and 2 (right).}
    \label{fig:eval_pallets_in_zone}
\end{figure}

Note that neither the ground loading zones nor the truck have predefined pallet positions. Pallets can be placed anywhere within these areas, with the restriction that they must be accessible by the forklift.
Example situations for ground loading zones are shown in Figure~\ref{fig:eval_pallets_in_zone} from the view of the pallet detector camera on top of the lift mast. For unloading in Zone 1, all four pallets were arranged in a single row, whereas in Zone 2 they were arranged in two rows with two pallets in each row.

\begin{figure}
    \centering
    \includegraphics[width=0.62\linewidth]{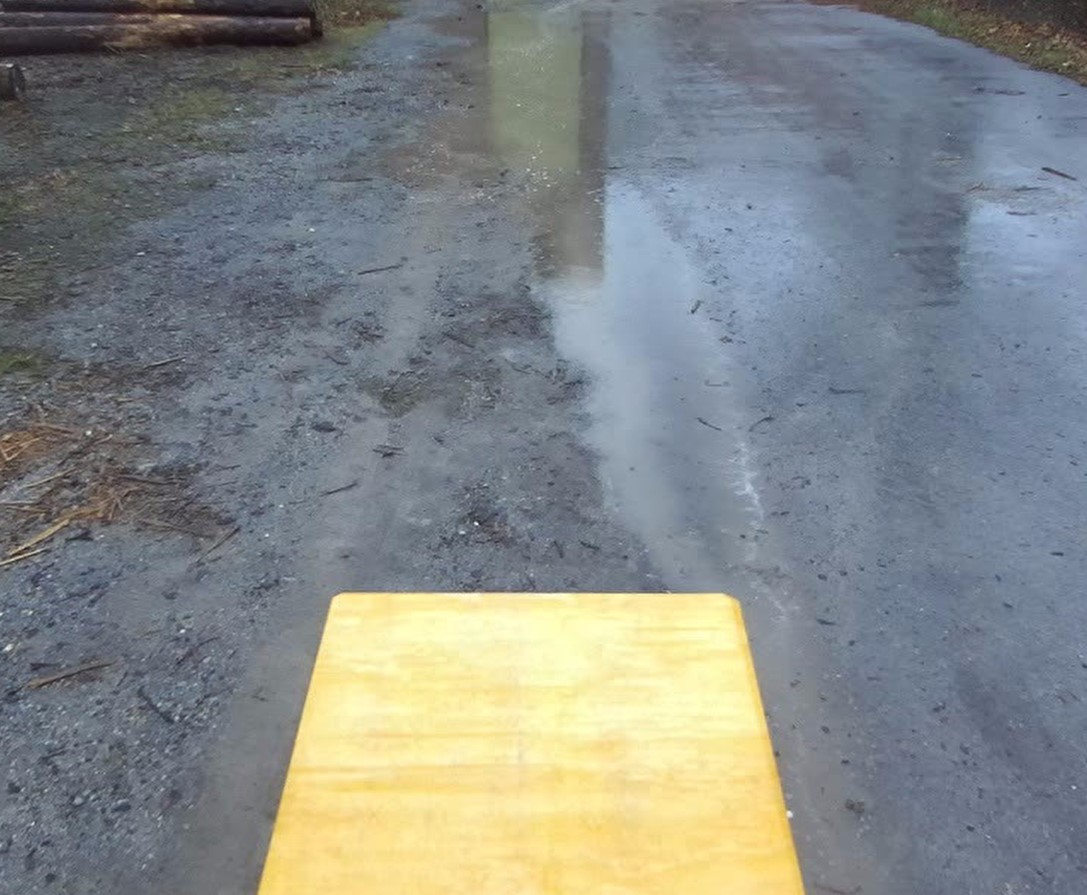}
    \caption{Rainy environment conditions for the ground-to-ground scenario with reflective puddles.}
    \label{fig:eval_rain_view}
\end{figure}

As noted above, one of the purposes of this work is to evaluate the autonomous system under varying weather conditions. Therefore, the baseline data, as well as the data from autonomous operation, are collected under different weather conditions, including sunny, cloudy, and rainy weather, with up to 2 mm/h rain intensity, as illustrated in \Fig{fig:eval_rain_view} for a ground-to-ground loading scenario. Table \ref{tab:eval_weather_conditions} summarizes the distribution of weather conditions for the evaluated data.

\begin{table}[ht]
    \centering
    \caption{Distribution of weather conditions for the evaluated data.}
    \label{tab:eval_weather_conditions}
    \begin{tabular}{c|c|c|c}
         Weather & Sunny & Cloudy & Rainy \\ \hline
         Ratio & 24\% & 44\% & 32\%
    \end{tabular}
\end{table}

\subsection{Baseline Data Collection}

To be able to assess the efficiency, accuracy and general performance of the autonomous forklift during typical loading scenarios, data from manual operation were collected.
The baseline data was collected through an experienced and external forklift operator with more than 20 years of experience in handling this type of forklift. 
The data collection process was supervised by an engineering team over a four days of testing, encompassing a variety of weather conditions. The study included multiple test scenarios, covering both loading and unloading from a truck as well as ground-level pallet handling. Each scenario was repeated across several cycles to generate statistically significant results. Overall, over 300 minutes of consecutive automated driving were collected.

\subsection{Limitations of the Experimental Evaluation}
\new{
All quantitative experiments were conducted within a single laboratory environment. This constraint was imposed deliberately: the primary objective of the evaluation was a controlled comparison against a human operator, which presupposes repeatable and reproducible conditions.
Within these controlled conditions, several sources of variability were retained. Trials were distributed across multiple days and times of day, introducing variation in ambient lighting and weather-related influences. While the loading zones and trucks remained at fixed positions, the pallet poses within each zone were not identical accoss trials, with an estimated variation of up to ±30 cm in translation and ±15° in orientation. In combination with the stochastic behaviour of the perception, localization, and motion-planning subsystems, this led to variation in the generated trajectories and the corresponding operational scenarios.
}

\new{
The experiments were further limited to Europallets, evaluated both without load and box-loaded configurations, with sizes ranging from approximately 15 cm to 60 cm. Consequently, the results primarily reflect the system’s performance for standardized pallet geometries and a bounded set of load configurations.
Beyond the quantitative trials, the system was additionally successfully tested at external sites, including the DigiTrans\footnote{https://www.digitrans.expert/en/outdoor-rain-plant/ (accessed 2026-06-03)} artificial rain facility, to examine its qualitative behaviour under more adverse environmental conditions.
Future experiments will consider a broader set of pallet types, load geometries, environmental conditions, and deployment sites to assess the generalizability of the proposed approach.
}

\subsection{Performance Criteria}

The manual operation was recorded with the same forklift to compare the performance of \ADAPT with that of an expert operator. The maximum velocity in autonomous operation was limited to the legally maximum allowed velocity of $v_{max}=6$km/h. The following performance criteria are selected for evaluation:

\textbf{Time taken for full loading cycle:} Total time taken to complete the entire pallet loading and unloading process, from the first pallet pick-up to the last pallet placement. This provides a direct measure of throughput, allowing a clear comparison of speed and productivity between the autonomous system and the human operator.

\textbf{Distance driven:} Total distance traveled for the full load task. This enables a comparison between the route efficiency of the autonomous system's algorithms and the decision-making of an expert operator. Lower travel distances also equal less energy consumption.

\textbf{Manual interventions:} Quantity and severity of manual interventions required during autonomous operation. The goal is to analyze the autonomy of the system and its need of human involvement to correct errors, manage obstacles, or ensure safety.

Using these criteria, it is possible to conduct a comprehensive, quantitative and qualitative comparison of \ADAPT and an expert operator, with an emphasis on efficiency and operational performance.

\subsection{Performance Analysis}

\begin{figure}
    \centering
    \includegraphics[width=\linewidth]{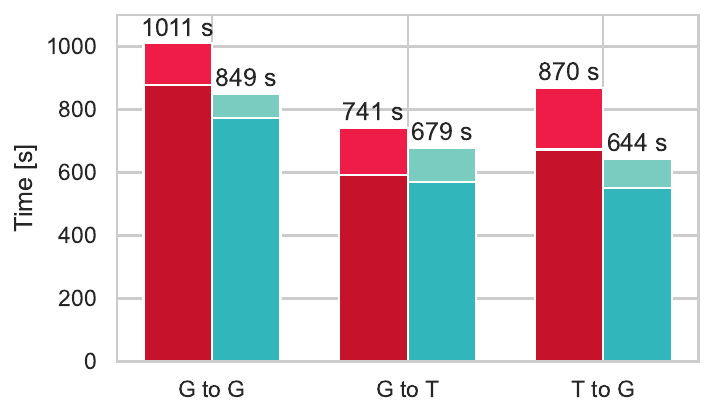}
    \vspace{-20pt}
    \caption{Overall mean time for the three scenarios ground-to-ground (G to G), ground-to-truck (G to T) and truck-to-ground (T to G) for autonomous operation (red) and the expert operator (cyan). Time spent driving in darker, standstill in lighter color.}
    \label{fig:eval_overall_time}
\end{figure}

\textbf{Time taken for full loading cycle:}
Figure~\ref{fig:eval_overall_time} compares the overall task completion times for autonomous and manual operations across three scenarios: ground-to-ground (G to G), ground-to-truck (G to T), and truck-to-ground (T to G). The figure further distinguishes between time spent driving (shown in a darker shade) and non-driving task execution (shown in a lighter shade). Overall, the experienced operator completed the loading tasks in 83\% of the time required by \ADAPT.

In the simpler ground-to-
ground scenario, the autonomous system achieves approximately 84\% of expert-level performance, demonstrating its strong potential for timely deployment in commercial operations. In the more complex ground-to-truck scenario, the autonomous system remains competitive, showing a time difference of 14\% compared to the performance of experts. The largest discrepancy, 26\%, is observed in the truck-to-ground scenario. The extended standstill times observed in \ADAPT primarily result from computational and detection delays, as well as inefficiencies associated with the sequential execution of tasks, as discussed in detail later.

\begin{figure}
    \centering
    \includegraphics[width=\linewidth]{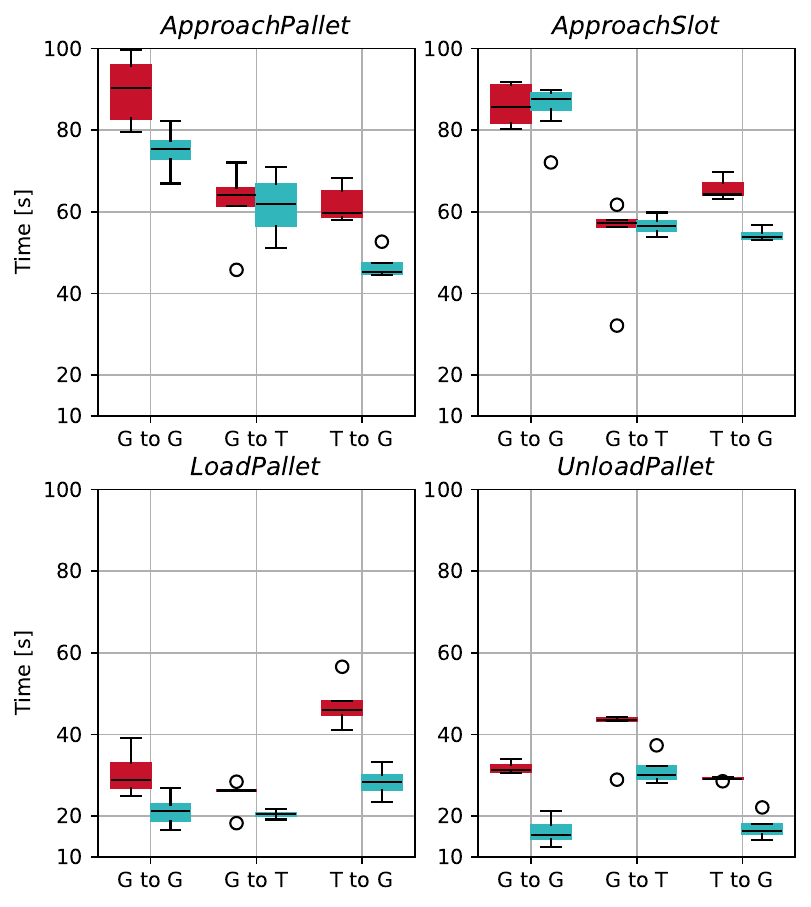}
    \vspace{-20pt}
    \caption{Time evaluation of the sub-missions for ground-to-ground (G to G), ground-to-truck (G to T), and truck-to-ground (T to G) loading for autonomous operation (red) and the forklift operator (cyan).}
    \label{fig:eval_boxplot_time}
\end{figure}

To evaluate the potential for performance and improvement in more detail, Figure~\ref{fig:eval_boxplot_time} provides a statistical time analysis based on the subtasks of autonomous operation. The missions discussed in Section~\ref{sec:planning_control} are \textit{ApproachPallet}, \textit{LoadPallet}, \textit{ApproachSlot}, and \textit{UnloadPallet}. The two submissions \textit{FindPallets} and \textit{SelectPallets} are included in \textit{ApproachPallet} to increase readability. The two \textit{Approach} missions consist mostly of typical navigation movement, whereas the load and unload missions deal mostly with manipulation.

\begin{figure*}[ht!]
    \centering
    \includegraphics[width=\linewidth]{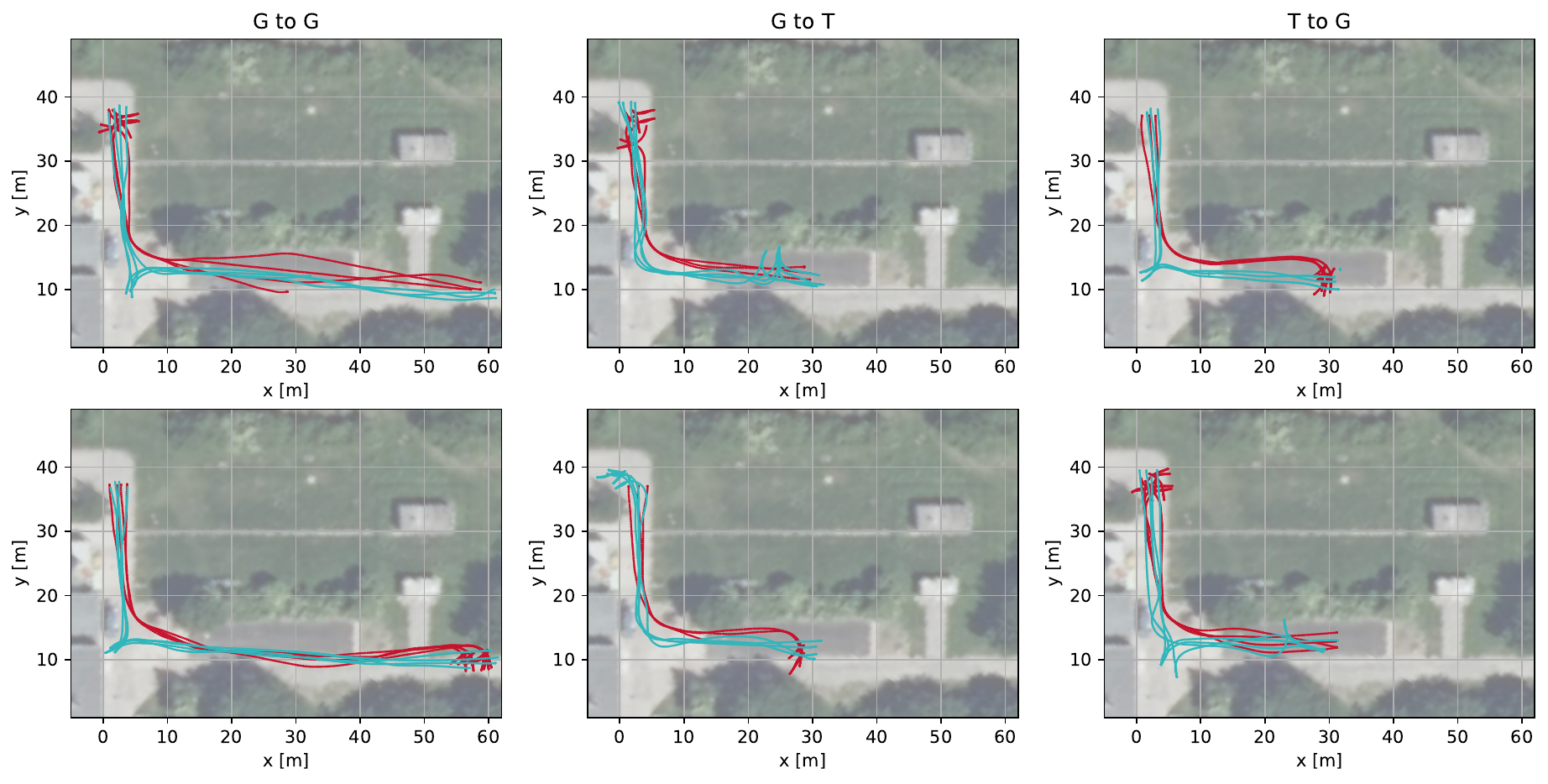}
    \vspace{-20pt}
    \caption{Paths for autonomous (red) and manual (cyan) operation for all 3 scenarios for \textit{ApproachPallet} (top) and \textit{ApproachSlot} (bottom).}
    \label{fig:eval_path_overlay}
\end{figure*}

For the \textit{ApproachPallet}, a longer time and a higher variance for \ADAPT can be observed. This arises mostly from two properties of the pallet detection: Firstly, when no pallets are known to the system, \ADAPT stops for several seconds in front of the loading zone to obtain a stable object list. Secondly, as the position and orientation of the pallets are not known in advance, a new approach of pallets is sometimes mandatory for correct loading which takes up to 30 seconds. This occurred multiple times, especially for the ground-to-ground scenario, resulting in a larger variance for the approach. In the ground-to-truck case, the performance of \ADAPT and the expert is very similar because the expert operator chooses a suboptimal path, as detailed in the latter. For the \textit{ApproachSlot} mission, \ADAPT achieves a very similar performance, even though more changes in driving direction were made. 

For the manipulation subtasks, \ADAPT performs very consistently for all scenarios. For loading, the performance of ground-to-ground and ground-to-truck is very similar to that of the human operator. The loading of the truck takes longer for \ADAPT. This mostly arises from the time needed to raise the forks to the height of the loading platform and the additional pallet detection check when the pallet is located in a plane different from the movement plane. The unloading of the pallets is close to a standardized task and thus can be done with accurate timing with hardly any variance by the autonomous system. The key difference from manual operation is that the autonomous system raises the forks while stationary, as shown in Figure~\ref{fig:pallet_putdown}, while a human operator simultaneously manipulates the forks and drives to maximize efficiency. This illustrates a case where \ADAPT is not yet optimized for timing performance.

\begin{figure}[!]
    \centering
    \includegraphics[width=1.0\linewidth]{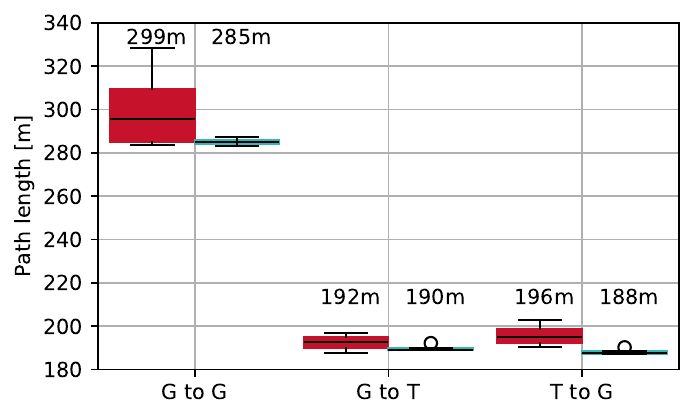}
    \vspace{-10pt}
    \caption{Path length evaluation for autonomous (red) and manual (cyan) operation. Text denotes the absolute path length.}
    \label{fig:eval_path_length}
\end{figure}

\textbf{Distance driven:}
Figure~\ref{fig:eval_path_overlay} illustrates an overlay of the manual and autonomous paths for a representative loading cycle in each scenario. In general, it can be observed that the expert operator shows a superior utilization of the free space available for turning compared to the autonomous planning system. This results in fewer changes in driving direction during manual operation. In contrast, the autonomous vehicle tends to navigate closer to obstacles, such as bushes, allowing it to effectively cut corners. These opposing behaviors nearly compensate for each other, resulting in comparable path lengths, as summarized in Figure~\ref{fig:eval_path_length}. Contrary to our intuition, the expert operator performs impressively consistent with a very low variance of the path length. For \ADAPT, the variance in path length is greater as a result of the heuristic nature of the hybrid A* path planner. This can also be seen in the plotted paths in Figure~\ref{fig:eval_path_overlay}, where the paths of the human operator are impressively equal within one scenario. Additionally, in one instance, only three of the four pallets were loaded autonomously, prompting a human intervention. \ADAPT then restarted from its home position and loaded the remaining pallet autonomously, which resulted in an additional trip from the home position to the loading area, producing a significant outlier in the G to G scenario.

Across different scenarios, a notable change in human operator behavior was observed, as Figure~\ref{fig:eval_path_overlay} shows. After the first recorded scenario, ground-to-truck loading, the expert operator adapted his turning strategy. Instead of turning immediately after loading, the operator used the corner as a natural pivot point to change the driving direction. This adjustment was shown to be significantly faster, as evidenced by the reduced driving time in the \textit{ApproachPallet} phase compared to the truck-to-ground scenario. These observations highlight the dependency of performance on the skill level of the forklift operator. Less skilled operators not only take longer to load and unload pallets but also tend to select suboptimal driving paths, further impacting overall efficiency.

\begin{figure}
    \centering
    \includegraphics[width=\linewidth]{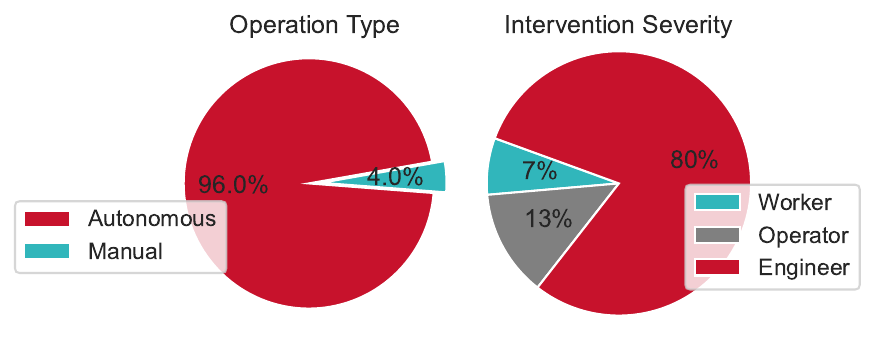}
    \caption{Left: Time spent on autonomous operation (red) compared to manual interventions (cyan). Right: Time spent for manual intervention categories, grouped by severity.}
    \label{fig:eval_manual_intervention_time}
\end{figure}

\textbf{Manual interventions:}
During 303 minutes of autonomous operation, 84 pallets were loaded and only 13 minutes were spent in manual control, accounting for  4\% of the total operation time. Most of this manual intervention, 10 minutes, was dedicated to resolving a GNSS localization issue, which is primarily a prototype-related challenge. All other combined manual interventions accounted for just 3 minutes, representing less than 1\% of the total operation time. \Fig{fig:eval_manual_intervention_time} presents the time analysis of manual interventions during autonomous operation. 

To analyze the manual interventions in more detail, three  categories of expertise for resolving are introduced: 1) \textit{Worker} for a typical worker on-site with little knowledge of the machine; 2) \textit{Operator}, denoting an educated forklift operator capable of driving the machine manually; and 3) \textit{Engineer}, as a robotic engineer with in-depth knowledge of the autonomy framework. The right-hand side of \Fig{fig:eval_manual_intervention_time} shows that most of the time was needed for the \textit{Engineer} level, \ie the above mentioned 10 minutes.

In absolute numbers, a total of 25 manual interventions were required, which amounts to under 5 interventions per hour. Table \ref{tab:eval_man_int_for_weather} shows the distribution of manual interventions over the weather conditions. Notably, sunny weather required the most interventions. These interventions occurred because of dust triggering the collision avoidance module, as well as lower detection quality of the pallets under direct sunlight. For rainy conditions, the best performance with respect to the manual interventions was recorded. This marks the good quality of the stereo-camera detection under rain as long as the camera lens is protected from direct raindrops.

\begin{table}[h]
    \centering    
    \caption{Manual interventions per hour under different weather conditions.}
    \label{tab:eval_man_int_for_weather}
    \begin{tabular}{c|c|c|c|c}
         & total & sunny & cloudy & rainy \\ \hline
         Manual interventions / h & 4.8 & 5.9 & 5.8 & 2.5 
    \end{tabular}
\end{table}

\begin{figure}
    \centering
    \includegraphics[width=\linewidth]{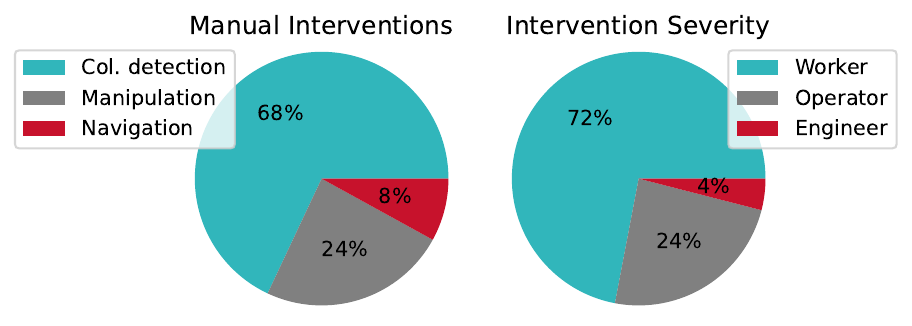}
    \caption{Overall manual interventions for the autonomous loading operation categorized into type and severity.}
    \label{fig:eval_manual_interventions}
\end{figure}

Further, the manual interventions were classified into Collision Detection, Manipulation, and Navigation. The distribution of types of manual intervention is illustrated on the left-hand side of Figure~\ref{fig:eval_manual_interventions}. Notably, nearly over two-thirds of these interventions were attributed to the collision avoidance module, which requires human clearance for any detected potential collision. Many of these collision warnings occurred near the boundary of the map, where the distance from the collision is shorter if approaching an obstacle or boundary of the map. 
This issue, resulting from stringent safety constraints in the laboratory environment, can be alleviated by implementing obstacle-classification-based warning distances, assigning larger safety margins for humans and moving obstacles, and smaller margins for static obstacles. Additionally, collision warnings were correctly triggered when humans, including the safety operator, were detected in the vehicle's path, although such incidents were rare. The second category involves inaccuracies in the detection and manipulation modules, which result in failed pallet load and unload attempts. Implementing recovery routines could reduce the number of necessary manual interventions, albeit at the cost of longer execution times. Lastly, navigation and localization interventions occur when no path to a desired position can be planned, or the localization module fails.

On the right-hand side of Figure~\ref{fig:eval_manual_interventions}, the severity of manual interventions is classified. Contrary to the time requirement, most human interventions can be solved by a regular worker, commonly available at construction sites. For the collision avoidance, simply ensuring that the corridor in front of the autonomous vehicle is free of obstacles is sufficient. For \ADAPT, this is verified by pressing a confirmation button once.
Approximately one-quarter of interventions require an operator for forklift operations to assist in navigation or manipulation.
In our tests, there was one instance where an engineer with knowledge of autonomous functions was needed, due to a failure in the GNSS localization system.
These results highlight the substantial reduction in manual labor required for material transport, demonstrating the efficiency of \ADAPT.
Table \ref{tab:eval_summary} summarizes the evaluated key performance indicators compared to the human expert and the ratio of them, where $\eta>100\%$ marks superior autonomous operation.
\begin{table}[h]
    \centering
    \caption{Summary of key performance indicators for \ADAPT, the human expert, and the ratio $\eta$ = Expert / \ADAPT.}
    \label{tab:eval_summary}
    \begin{tabular}{c|c|c|c}
         KPI &  \ADAPT & Expert & $\eta$ \\ \hline
         Mean loading time & 14.6 min & 12.1 min & 83\% \\         
         Mean autonomous operation & 14.0 min & 0.0 min & - \\
         Mean manual operation & 0.6 min & 12.1 min & 2017\%\\
         Mean path length & 225 m & 221 m & 98\% \\
         Manual interventions & 4.8 / h & - & - \\
    \end{tabular}
\end{table}

\section{Conclusion and Outlook}
\label{sec:conclusion}
This article presented the autonomous outdoor forklift ADAPT, a system capable of operating in the complex and unstructured environments of construction sites, even in challenging conditions, including low to medium rain. By integrating a forklift-specific sensor suite with a tailored software stack, the system achieves fully autonomous pallet loading and transportation. Extensive real-world testing demonstrated that the autonomous forklift operates at near-human efficiency, achieving more than 80\% of expert-level performance in the demonstrated scenarios, while requiring minimal operator intervention.

A key contribution of this work is the novel factor-graph-based joint optimization approach for vehicle localization and pallet mapping, specifically tailored for object manipulation tasks such as pallet loading. This method enhances the system’s adaptability and accuracy in dynamic construction site environments. Additionally integrating a novel fork contact measurement based on pressure feedback significantly improves robustness and safety during object manipulation while remaining cost-effective and simple to implement.

A detailed evaluation compared the autono\-mous forklift’s performance to that of an expert human operator with over 20 years of experience. The analysis provided insights into performance, robustness, and the nature and frequency of required human interventions. These findings highlight the system’s potential to improve efficiency and safety in material handling for construction sites, addressing labor shortages and reducing operational risks.

Future improvements will focus on motion planning in dynamic environments to enhance safety while increasing operational efficiency by reducing the number of manual interventions. 
Additionally, the incorporation of semantic information into the environment mapping approach will complement the existing geometric model,
allowing the forklift to prioritize stable surfaces such as concrete roads over more challenging terrains like gravel or soil. 
Further efforts will be directed toward increasing robustness when navigating through challenging terrain and extreme weather conditions.
Research on generalizing the system to different pallet types is ongoing and will support adaptation to a wider range of industrial environments. 
Recurring hardware updates, particularly those involving processing components, will allow replacing existing hardware with more compact, resource-efficient alternatives or utilizing higher-performance components to improve overall timing-related performance. 
Collectively, these improvements will advance the system toward a fully integrated, intelligent solution for autonomous material handling, adaptable to a range of industrial settings, such as demanding construction environments and intralogistic yards.

\section*{Conflicts of Interest}

The authors declare no conflicts of interest.

\section*{Data Availability Statement}

The data that support the findings of this study are available from the corresponding author upon reasonable request.

\section*{Acknowledgments}
This work was funded by the European Union’s Horizon 2020 research and innovation program under Grant Agreement No. 101006817 and supported by the Austrian Research Promotion Agency (FFG) through the ICT of the Future program under Project No. 873987.
We gratefully acknowledge Florian Wimmer for the initial version of the loading platform detector. 
We further extend our gratitude to our collaboration partners FH OÖ, Wolfgang Pointner, and Palfinger for their continued cooperation and support, as well as Austrian Scientific Computing (ASC) for providing large-scale computing resources and technical support.

\bibliography{
bibtex/bib/manual_literature_ordered
}

\clearpage  
\onecolumn
\FloatBarrier       
\appendix
\section{Behavior Tree}
\label{app:behavior_tree}

\begin{figure}[ht]
    \centering
    \includegraphics[angle=-90,width=0.33\linewidth]{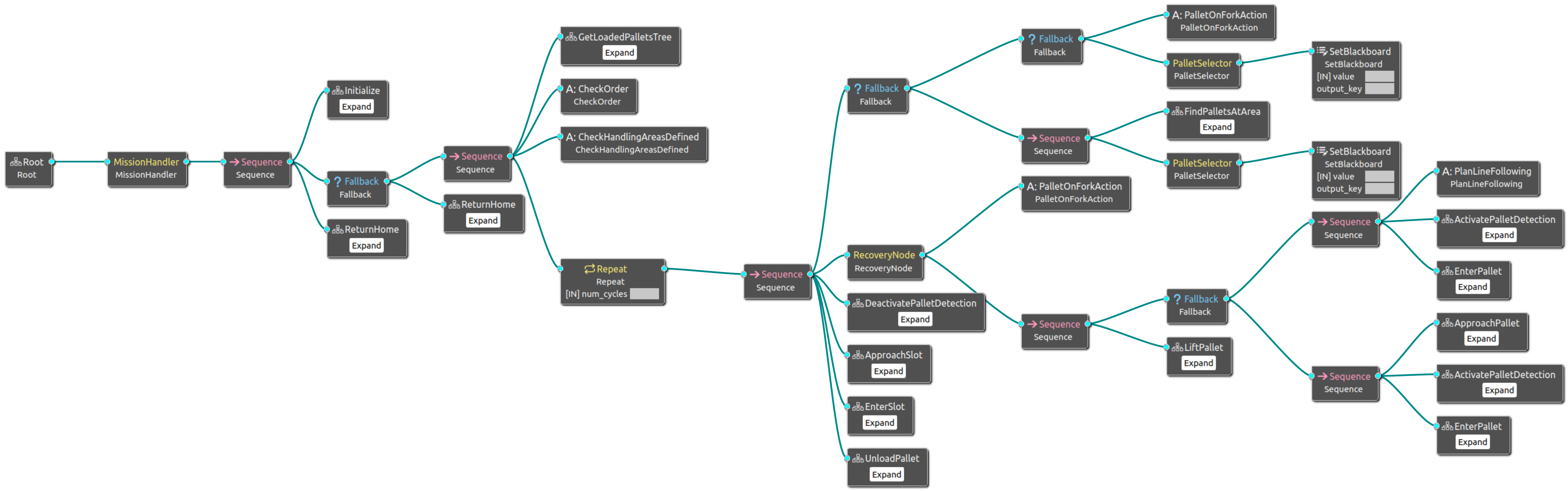}
    \caption{Main behavior tree designed for the proposed system and the experiments in Section~\ref{sec:experiments}.}
    \label{fig:loading_bt}
\end{figure}

Figure \ref{fig:loading_bt} illustrates the main behavior tree, which defines the structure and flow of tasks used in the experiments described in Section~\ref{sec:experiments}. It serves as a reference for understanding the system's decision-making process and task execution. The tree comprises 26 nested sub-trees and more than 30 unique action and composite nodes.

\end{document}